\documentclass[11pt]{article}

\usepackage[final]{acl}

\usepackage{times}
\usepackage{latexsym}

\usepackage[T1]{fontenc}

\usepackage[utf8]{inputenc}

\usepackage{microtype}

\usepackage{inconsolata}

\usepackage{graphicx}

\usepackage{amsmath,amsfonts,bm}

\def\eqref#1{equation~\ref{#1}}

\def\1{\bm{1}}

\DeclareMathAlphabet{\mathsfit}{\encodingdefault}{\sfdefault}{m}{sl}
\SetMathAlphabet{\mathsfit}{bold}{\encodingdefault}{\sfdefault}{bx}{n}

\newcommand{\R}{\mathbb{R}}

\usepackage[utf8]{inputenc} 
\usepackage[T1]{fontenc}    
\usepackage{url}            
\usepackage{booktabs}       
\usepackage{amsfonts}       
\usepackage{nicefrac}       
\usepackage{microtype}      
\usepackage{adjustbox}

\usepackage{times}
\usepackage{latexsym}
\usepackage{caption}
\usepackage{multirow} 
\usepackage{makecell}
\usepackage{graphicx}
\usepackage[table]{xcolor}
\usepackage{subcaption}
\usepackage{amsmath}
\usepackage{siunitx}
\usepackage{vcell}
\usepackage{amssymb}

\usepackage{hyperref}
\usepackage{cleveref}
\usepackage{arydshln}
\usepackage{scalerel}
\usepackage{xparse}
\usepackage[most]{tcolorbox}
\usepackage{enumitem}

\newcommand{\todo}[1]{}
\renewcommand{\todo}[1]{{\color{red} TODO{#1}}}
\newcommand{\benchmark}{\textbf{\textsc{RExBench}}}

\NewDocumentCommand\rexmoji{}{\scalerel*{\includegraphics{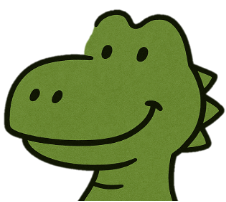}}{X}}

\title{\rexmoji{} \benchmark{}: Can coding agents autonomously \\ implement AI research extensions?}

\author{
\textbf{Nicholas Edwards\textsuperscript{1,2,*}}\quad
\textbf{Yukyung Lee\textsuperscript{3,*}}\quad
\textbf{Yujun (Audrey) Mao\textsuperscript{3}}\quad
\textbf{Yulu Qin\textsuperscript{3}}\\
\textbf{Sebastian Schuster\textsuperscript{1,$\dagger$}}\quad
\textbf{Najoung Kim\textsuperscript{3,$\dagger$}}
\\[0.5em]
\textsuperscript{1}Faculty of Computer Science, University of Vienna, Vienna, Austria\\
\textsuperscript{2}UniVie Doctoral School Computer Science, University of Vienna, Vienna, Austria\\
\textsuperscript{3}Boston University
\\[0.5em]
{\small \texttt{\{nicholas.edwards, sebastian.schuster\}@univie.ac.at}} \\
{\small \texttt{\{ylee5, amao, yuluqin, najoung\}@bu.edu}}
\\}
\begin{document}
\maketitle

\begingroup
\renewcommand\thefootnote{}\footnotetext{
\textsuperscript{*}Equal contribution. The order of co-first authors was randomly determined.
\textsuperscript{$\dagger$}Corresponding authors.}
\endgroup

\begin{abstract}
Agents based on Large Language Models (LLMs) have shown promise for performing sophisticated software engineering tasks autonomously. In addition, there has been progress towards developing agents that can perform parts of the research pipeline in machine learning and the natural sciences. We argue that research \textit{extension} and its implementation is a critical capability for such systems, and introduce \benchmark{} to support the evaluation of this capability. \benchmark{} is a benchmark consisting of realistic extensions of 12 research papers that aim to investigate \textit{novel} research hypotheses. Each task is set up as an extension to an existing research paper and codebase, accompanied by domain expert-written instructions. \benchmark{} is robust to data contamination, and supports an automatic evaluation infrastructure that executes agent outputs to determine whether the success criteria are met. We use this benchmark to evaluate 12 LLM agents implemented using two different frameworks: aider and OpenHands. We find that all agents fail to autonomously implement the majority of the extensions, with the best agent at around 33\% success rate. Although the success rate improves with additional human-written hints, the best performance under this setting remains below 44\%. This indicates that current agents are still short of being able to handle realistic research extension tasks without substantial human guidance. 

\noindent\centering
\begin{adjustbox}{width=\linewidth}
\begin{tabular}{@{}c@{\hspace{6pt}}l@{}}
  \raisebox{-0.25\height}{\includegraphics[height=11pt]{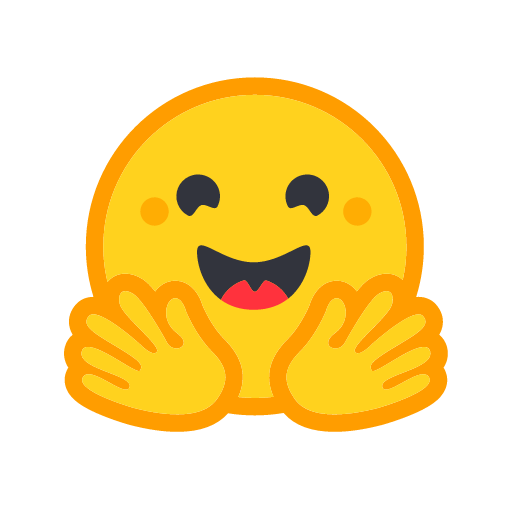}} &
  \href{https://huggingface.co/datasets/tin-lab/RExBench}{\texttt{huggingface.co/datasets/tin-lab/RExBench}} \\
  \raisebox{-0.25\height}{\includegraphics[height=11pt]{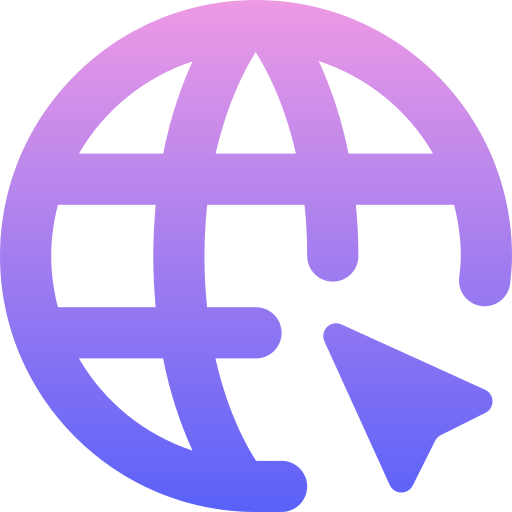}} &
  \href{https://rexbench.com}{\texttt{https://rexbench.com}} \\
\end{tabular}
\end{adjustbox}
\end{abstract}

\section{Introduction}
Interesting research necessarily builds on other research. In this regard, \textit{extensions} of existing research are important starting points to new investigations, potentially building up towards exciting novel discoveries. In light of recent growing interest in building Large Language Model (LLM) agents that can conduct scientific research in an autonomous manner, we propose \benchmark{}, a benchmark aiming to evaluate LLM agents' ability to extend existing AI research, with an initial focus on Natural Language Processing (NLP) and Machine Learning (ML). More specifically, \benchmark{} tests whether LLM agents can autonomously implement research extension experiments via code in a hypothesis-guided manner \citep{luo2025llm4sr}, where the extension hypotheses are provided to the system as verbal instructions along with relevant background material including the research paper(s) and the corresponding codebase. Our benchmark consists of realistic extensions of 12 recently published research papers in the field, accompanied by domain expert-written extension instructions (see Appendix~\ref{appendix:winodict_instructions} for a sample task instruction). The extension tasks cover various aspects of implementation involving changes to the model, algorithm, data, and evaluation method. The main metric of success is numerical replication of the outcome of domain-expert implemented ``gold'' solutions for the extension task. We provide an automatic evaluation infrastructure to execute the LLM agent-implemented solutions and evaluate the outcomes. The executions of both the gold solutions and system solutions are conducted in virtual machines with exactly the same specifications to control for experimental variation. \benchmark{} furthermore is robust to data contamination issues that affect the majority of existing benchmarks: the solutions and the success criteria for our extension tasks only exist in our held-out evaluation infrastructure and do not exist anywhere online.

\begin{figure*}[t]
    \centering
    \includegraphics[width=\linewidth]{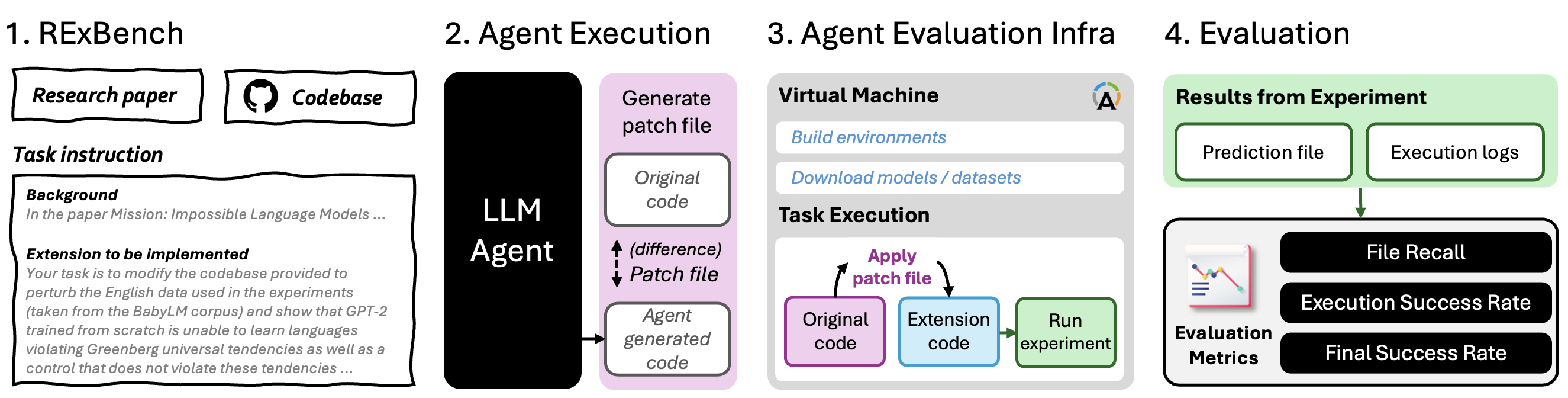}
    \caption{End-to-end workflow of \benchmark{}: (1) An LLM agent receives inputs consisting of the research paper(s), the original codebase, and an extension instruction; (2) the system implements the extension and a patch file is obtained; (3) the patch is applied to the original code and executed via our evaluation infrastructure; and (4) the results are evaluated using specified metrics.}
    \label{fig:main_figure}
\end{figure*}

We tested twelve agents based on an array of LLM backbones (Claude 4/3.7 Sonnet \citep{anthropic2025claude4sonnet, anthropic2024claude37sonnet}, GPT-5 \citep{openai2025gpt5}, o1 \citep{jaech2024openai}, o4-mini, and DeepSeek-R1 \citep{guo2025deepseek}, using two different agent frameworks (aider and OpenHands). Many agents struggled on our benchmark, achieving success rates close to zero for most tasks. Agents with Claude 4/3.7 Sonnet and GPT-5 as backbone showed promise, often showing qualitative signs of success even when they did not achieve final success. Nevertheless, even the best-performing agent succeeded only about one third of the time on average (33\% success rate for OpenHands + Claude 4 Sonnet), leaving much headroom for progress.

While the current \benchmark{} tasks pose substantial challenges for the agents tested, most extensions do not require major rewriting of the codebase and are not extremely challenging in terms of complexity (at least to a PhD-level domain expert). We thus consider the release of this specific set of tasks and the paper as a contribution about the broader framework for evaluating research extensions (and the opportunities it may bring), which will motivate the development of more challenging extensions covering broader scientific domains, inviting contributions from the community.
\section{Related Work}

Recent advancements in LLMs and agentic frameworks have motivated discussions about their applicability to scientific research. This includes using LLMs and LLM agents for research automation \citep{li2025scilitllm, skarlinski2024language, jansen-etal-2025-codescientist, ziems2023can, choi2024use, boiko2023autonomous, gottweis2025towards, kitano2021nobel, gandhi2025boxinggym, mitchener2025kosmos} and benchmarking their ability to conduct research in the domains of social sciences, statistics, and natural sciences \citep{tian2024scicode, chen2025scienceagentbench, laurent2407lab, sharma2026researchrubrics}. For ML research, current attempts use agents to automate all stages of the research process: from ideation \citep{si2025can} to experiment design \citep{abramovich2025ablationbench, liu2025guide, zhao2025abgen} and execution \citep{siegel2024corebench, xiang2025scireplicatebench, miao2026recodeh}, to paper review and meta review \citep{du2024llms}. There have also been early attempts to automate the full research pipeline \citep{lu2024ai,kon2025curie, miyai2026jr}. 

Another line of work benchmarks coding and software engineering skills. Specific skills targeted include improving iterative coding capabilities \citep{yang2025codeclash},  resolving GitHub issues \citep{jimenezswe, badertdinov2025swerebench, deng2025swe}, debugging LeetCode problems \citep{tian2024debugbench}, resolving configuration and dependency issues in research environment setups \citep{bogin2024super}, and solving tasks in a terminal environment \citep{merrill2026terminalbench}. In a similar vein, other benchmarks assess more comprehensive ML problem-solving and code implementation skills. MLE-bench \citep{chan2025mlebench} and DSBench \citep{jing2025dsbench} design machine learning and data science tasks akin to Kaggle-style competitions; MLAgentBench \citep{huang2024mlagentbench} gathers classical ML tasks, including Kaggle challenges; DataSciBench evaluates data analysis and visualization skills with novel evaluation pipelines \citep{zhang2025datascibench}; ML-Dev-Bench \citep{padigela2025ml} focuses on the full ML development workflow; and MLGym challenges agents' full pipeline research skills with tasks in domains such as computer vision and NLP \citep{nathani2025mlgym}. 

The most directly relevant efforts to ours are agentic frameworks and benchmarks that focus on ML problem-solving and software engineering capabilities in research settings. Curie \citep{kon2025curie} aims to evaluate the ability to plan and execute experiments; BLADE \citep{gu2024blade} is designed to automatically evaluate agents' approaches to open-ended data-driven research questions. ResearchCodeBench \citep{hua2025researchcodebench} tests LLMs' abilities to code ML ideas derived from recent research papers; Paper2Code \citep{seo2026papercode} introduces a multi-agent LLM framework to translate ML papers into codebases through a stage-wise design; PaperBench \citep{starace2025paperbench} evaluates research agents using a compilation of coding tasks targeting the replication of 20 ICML papers; and DeepCode \citep{li2025deepcode} proposes agentic frameworks that targets long-context coding challenges, such as PaperBench tasks. 


\benchmark{} has a similar goal to PaperBench and, to some extent, Curie, in benchmarking ML and AI research code generation. However, a key distinction is that instead of evaluating replications (PaperBench) or very general questions that can often also be answered without running experiments (Curie), we focus on \textit{novel research extensions}. Thus, \benchmark{} is able to evaluate agent performance on previously unseen or unimplemented research hypotheses, which greatly alleviates data contamination concerns.
\section{Benchmark Design}

\begin{table*}[t!]
  \centering
  \renewcommand{\arraystretch}{1.2}
  \resizebox{\linewidth}{!}{%
  \begin{tabular}{llp{7.5cm}l}
    \toprule
    \textbf{Identifier} & \textbf{Extension Type} & \textbf{Task Summary} & \textbf{Venue} \\
    \midrule
    \makecell[tl]{CheckEval\\\citep{lee-etal-2025-checkeval}} & Evaluation & Train a regression model to learn question-specific weights for LLM-as-a-judge outputs. & EMNLP 2025 \\
    \midrule
    
    \makecell[tl]{COGS\\\citep{kim2020cogs};\\\citep{csordas2021devil}} & Model & Retrain and evaluate the model without early stopping. & \makecell[tl]{EMNLP 2020;\\EMNLP 2021} \\
    \midrule
    
    \makecell[tl]{Entity Tracking\\\citep{kim2024code}} & Model & Evaluate a multi-modal model (Llama-3.2-11B-Vision). & Preprint \\
    \midrule
    
    \makecell[tl]{Explain then Translate\\\citep{tang2023explain}} & Algorithm & Assess problem complexity using cyclomatic complexity. & EMNLP Findings 2023 \\
    \midrule
    
    \makecell[tl]{Instruction Tuning\\\citep{hewitt2024instruction}} & Model & Extend the implementation for the rule-based instruction-tuning experiment to OLMo-7B. & Preprint \\
    \midrule
    
    \makecell[tl]{Mission Impossible\\\citep{kallini2024mission}} & Data/Evaluation & Compare learning on standard English vs. data perturbed by unattested linguistic constraints. & ACL 2024 \\
    \midrule
    
    \makecell[tl]{Othello\\\citep{li2023emergent};\\\citep{nanda2023emergent}} & Data/Evaluation & Implement a different representation of the game state in the probe (current vs. other player instead of white vs. black player). & \makecell[tl]{ICLR 2023;\\BlackboxNLP 2023} \\
    \midrule
    
    \makecell[tl]{Reasoning or Reciting\\\citep{wu2024reasoning}} & Model & Replicate the results using an open-source model (Llama-3.1-8B-Instruct). & NAACL 2024 \\
    \midrule
    
    \makecell[tl]{Re-reading\\\citep{xu2024re}} & Algorithm & Test strategy on Big-Bench Hard multi-step arithmetic using Llama-3.1-8B-Instruct. & EMNLP 2024 \\
    \midrule
    
    \makecell[tl]{Tree of Thoughts\\\citep{yao2023tree}} & Algorithm & Investigate the failure mode of the Tree of Thoughts algorithm on DeepSeek-V2-Lite-Chat. & NeurIPS 2023 \\
    \midrule
    
    \makecell[tl]{VariErr-NLI\\\citep{weber2024varierr}} & Model/Data & Train a distilled model to categorize annotations. & ACL 2024 \\
    \midrule
    
    \makecell[tl]{WinoDict\\\citep{eisenschlos2022winodict}} & Data/Evaluation & Replace synthetic target words with existing English words sampled at different frequencies. & EACL 2023 \\
    \bottomrule
  \end{tabular}}
  \caption{List of papers that form the bases for extensions in \benchmark{}.}
  \label{tab:papers}
\end{table*}

\subsection{Research Extension Task}
\label{sec:task-definition}

\paragraph{Task}
We define our research extension task as a code implementation problem, where the input consists of an existing research paper, an accompanying codebase, and an instruction that verbally describes an extension proposal and how this should be tested. To illustrate the level of complexity of \benchmark~tasks, consider the extension for WinoDict \citep{eisenschlos2022winodict}. The original work tested if LLMs can learn novel synthetic target words during inference from in-context dictionary definitions using Winograd-style co-reference resolution problems. A possible extension to this work is to assess whether using existing English words instead of novel words as the target words interferes with in-context word acquisition, where the hypothesis is that assigning new meanings to existing words is more difficult. Additionally, the frequency of the existing words may also modulate the learning effect. Therefore, in our setup the agent is instructed to modify the original paper's codebase to generate new datasets that replace the surface forms of the target words with existing English words sampled from specified frequency groups, where the correct inflected form of the word should also be generated. The full instruction for this example is provided in \Cref{appendix:winodict_instructions}. Given this input, a system must produce as output edits to the input codebase that implements the extension proposal and generates the new experimental results.

\paragraph{Desiderata}
The core aim of our benchmark is to \textit{automatically} assess how well an agent can \textit{autonomously} implement \textit{realistic} research extensions. These goals are to some extent in conflict with each other. Realistic research extensions tend to be quite open-ended, which makes automatic assessment challenging or impossible. On the other hand, limiting tasks to ones that can be scored by simple automatic measures may constrain the task too much for it to be still realistic. We strike a balance between these two goals by using automatic tests that allow the agent to tackle the task through any means, as long as this leads to results comparable to the ones from our gold implementation. The task setting of requiring implementation on top of an existing codebase and evaluation through controlled execution environments (random seed, hardware, packages, etc.) serves to improve the reliability of the numeric output-based automatic evaluation. Nevertheless, each extension proposal included in the benchmark still cannot be too open-ended or exploratory, and therefore consist of specifically-scoped questions that can have well-defined numeric targets. To ensure that agents autonomously implement extensions, the granularity of our instructions are calibrated at a level that still requires the agent to thoroughly analyze the codebase and form its own plan for the extension. Furthermore, at no point of the evaluation do humans provide additional supervision. Finally, one of the biggest challenges with LLM evaluation is data contamination. If solutions to any of the tasks are openly available on the web, LLMs that serve as the backbone for the agents may have been trained on the solutions (also noted as a possible issue in PaperBench: \citealt{starace2025paperbench}), rendering it impossible to establish whether success stems from memorization or autonomously solving the task. We circumvent this problem by including only novel research extensions, either in terms of the idea itself or implementation. To the best of our knowledge, none of our extensions exist on top of the existing codebases publicly; we store all the gold extensions in private Bitbucket repositories.\footnote{We use Bitbucket instead of GitHub since GitHub data has been used in the past to train LLMs and it is unclear whether this may also be true for some private repositories.} Furthermore, our privately hosted evaluation infrastructure prevents agents from accessing the evaluation scripts or reference solutions.

\subsection{Benchmark Composition}
Our benchmark consists of research extensions building upon papers and codebases primarily in the NLP and broader AI domains, taking into consideration the availability of expertise within the team as well as the availability/replicability of the code released. The full list of papers is in \Cref{tab:papers}.\footnote{Two of the tasks (COGS, Othello) involve implementing an extension proposal from another paper on top of the codebase of the original paper, where the implementation of the extension is either not publicly available or is not implemented as an edit of the original codebase. For these tasks, there are two relevant papers.} The specific extension proposals were selected to span various dimensions of change including changes to the model, dataset, algorithm, and evaluation. In addition to this consideration, we imposed the following constraints on the extension proposals for scientific rigor and feasibility of the experiments: (1) important empirical trends from the original paper relevant to the extension proposal must replicate; (2) the gold implementation of the extension proposal must replicate (e.g., if the gold implementation requires making calls to a closed API-based model, this may not replicate in the future due to model deprecation); and (3) the estimated runtime of each gold implementation should be shorter than 12 hours on a single A100 GPU. The final dataset includes the extension instruction, target research papers in both .pdf and .md format (converted using \texttt{PyMuPDF4LLM} to accommodate agents that lack the ability to read .pdf files), and the original codebase.

\subsection{Benchmark Construction Process}
For each extension proposal, a domain expert (PhD student-level or above) first verified that the original codebase replicates the results of the associated paper on our virtual machines (details to follow). Then, they implemented the ``gold'' edits for the target extension and recorded the numerical outcomes, ensuring that the runtime does not exceed 12 hours. This implementation process and the outcomes were validated by at least one other author. Finally, the domain expert wrote the instruction that consists of a brief description of the original paper, the extension proposal, and how this proposal should be tested (see \Cref{appendix:winodict_instructions} for an example of a full instruction). The description of the ``how'' was deliberately high-level to meet the desideratum of evaluating a sufficiently autonomous capacity. Nevertheless, since the instructions should not be confusing or ambiguous, they were polished through multiple rounds of revisions by multiple authors to improve clarity. Importantly, if we foresaw degrees of implementation freedom that may introduce random variation, we controlled for this by specifying constraints (e.g., use an implementation of Pearson correlation function from the \texttt{scipy} package as opposed to implementing this from scratch). During this revision process we furthermore ensured that each extension was self-contained. No part of the gold edits required information external to the set of inputs provided to the system. As a part of the revisions for self-containment, we provided information such as specific model identifiers and explanations of necessary hyperparameters not in any README or the paper as a part of the instruction, and added version information for all of the packages (via an \texttt{environment.yml} file).

\subsection{Evaluation Metrics}
\label{subsec:evaluation-metrics}
Our main metric is final success rate, which measures whether the outcome of executing the agent-implemented code falls within the target range. We define two additional metrics for finer-grained analyses: execution success rate and file recall. We describe each metric below.
\paragraph{Final Success Rate}
Final success rate evaluates whether the agent correctly implements the specified research extension. This evaluation either checks whether the final results exactly match the results of the gold implementation (if the run is fully deterministic) or fall within a \textit{gold range} (for runs with output variability). In the latter case, an agent solution is considered successful if its final execution outcome falls within this bound. For extensions with run output variability, we compute this bound by executing the gold implementation five times with different random seeds, setting the range to be $\pm$ 2 standard deviations from the mean result. In practice, these ranges are very narrow, and we furthermore observed no false positives (an incorrect agent implementation scoring within the range) or false negatives (a correct agent implementation scoring outside the range) during our evaluation.

\paragraph{Execution Success Rate}
Execution success rate checks whether the generated code runs without errors in our evaluation environment. This metric evaluates the general well-formedness of the code and contextual understanding sufficient to avoid runtime errors.
\paragraph{File Recall}
File recall quantifies whether files edited in the gold solution were also edited by the agent: 
$\text{File Recall} = |\text{Files}_{\text{agent}} \cap \text{Files}_{\text{gold}}| \,/\, |\text{Files}_{\text{gold}}|$.
The limitation of this measure is the dependency on the gold solution. Technically, a solution could achieve zero file recall with perfect final success. E.g., if an agent solution was exactly equivalent to gold but created new files with identical content instead of editing, and changed references appropriately in the repository, this would be the case. Still, we take human expert edits to reflect a reasonably efficient set of modifications. 
\subsection{Evaluation Infrastructure}
\label{subsec:eval-infra}
\paragraph{Submission format} Our metrics defined above require execution of agent generated code. We conduct this execution on a virtual machine to control for hardware specification and package dependencies. We host this infrastructure using our own resources to encourage community participation without resource concerns, and conduct evaluation asynchronously at a regular interval to update the leaderboard with the submissions we receive, similarly to \citet{jimenezswe}. The submissions are received in the form of git patch files (as opposed to full edited repositories) to streamline the submission process. We also request agent log files to verify that the task was completed autonomously by an agent.

\paragraph{Infrastructure pipeline} We host our evaluation infrastructure based on the OpenStack platform on an academic cloud computing service. For each patch file received, we execute the code in in a task-specific Apptainer container \citep{singularity_2021} that has the original codebase and evaluation scripts pre-loaded and the environment set up. To control for random variation of the execution outcomes to the best of our effort, we (1) fix all random seeds in the codebase wherever possible, and (2) run the evaluations with exactly the same hardware configuration as our gold runs (see \Cref{app:experiment-setup}, \Cref{tab:task-vm-info}). Inside the container, we apply the patch file and execute the task. We limit the runtime to 12 hours, which is around twice the duration of the gold solution with the longest runtime among our extension tasks (see \Cref{tab:task-vm-info} for all estimated runtimes). Once task execution is complete or the attempt crashes, we extract result files and task execution logs. This setup ensures a fully containerized and task-level parallelizable evaluation infrastructure.
\begin{figure*}
    \centering
    \includegraphics[width=\linewidth]{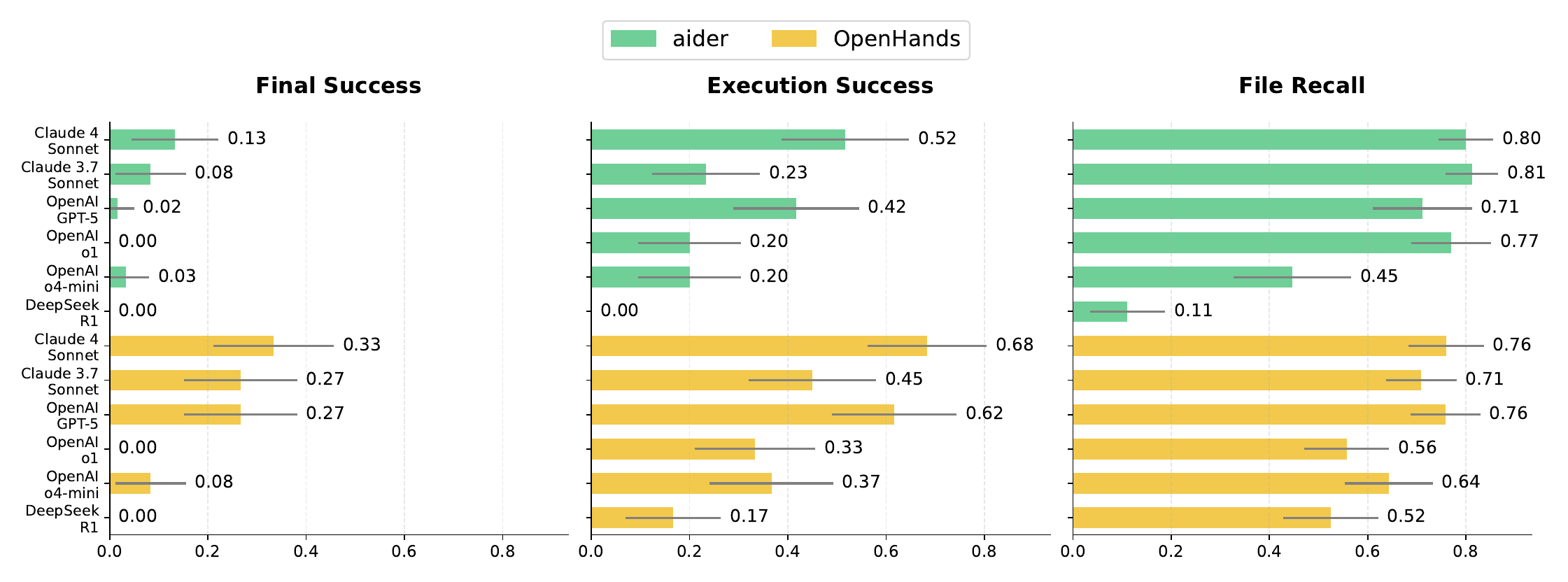}
    \caption{Agent performance on \benchmark{}. The color coding indicates the agent framework and the y-axis indicates the backbone LLM. Results include five runs per task to account for agent random variation. Error bars show standard error of the mean of all runs per model computed using the closed form formula (2$\sigma$, no normality assumption).}
    \label{fig:main_results}
\end{figure*}

\section{Experiments}
\subsection{Main Experiment}
We follow steps shown in \Cref{fig:main_figure} and evaluate twelve LLM agents, combining two agent frameworks with various LLM backbones (discussed below). We pass the full set of inputs for each task one by one to the agent to evaluate each task independently of each other. We run each task five times with the same agent model to account for agent random variation.
\subsubsection{Baseline agent design}
We used two different open-source agent frameworks (aider: \citealp{aider2023github} and OpenHands: \citealp{wang2024openhands}) that we adapted for the task. aider and OpenHands both support multiple backbone LLMs. We evaluated GPT-5, o1 and o4-mini (OpenAI), Claude 3.7 Sonnet and Claude 4 Sonnet (Anthropic), and an open-weight model (DeepSeek-R1). We discuss a few design decisions shared between our agents below. Note that this does not imply future submissions to our benchmark should be subject to the same design decisions.\footnote{After submission, we additionally evaluated Claude 4.5 Opus. See \Cref{appendix:additional_results_claude_4.5} for the results.}

\paragraph{Shared design considerations} For better runtime controllability, we disabled Python code execution for all agents. Regarding the settings of the backbone LLMs, we set the temperature to 0.7 for Claude 4/3.7 Sonnet and DeepSeek-R1. For GPT-5, o1, and o4-mini, we used the default settings, as these models do not support custom temperature adjustment. We specified the reasoning effort as medium for all OpenAI models. As discussed in \Cref{subsec:eval-infra}, our evaluation infrastructure requires git patch files. We created the patch files using a separate script after the agents had made changes to the codebase. We discuss individual implementation details in \Cref{app:agent-configurations}.\footnote{See our agent implementations at \href{https://github.com/tinlaboratory/RExBench-OpenHands}{OpenHands} and \href{https://github.com/tinlaboratory/RExBench-aider}{Aider}.}

\subsection{Experiment with Additional Hints}
We conduct an additional set of experiments where we provide different levels of hints to the agents. This experiment serves two purposes: (1) as a check that our tasks are possible to solve; (2) to diagnose where the difficulties lie, if the agents do find the tasks difficult without hints. We design two levels of hints, where the first level of hints provides help with information localization, and the second level of hints provides a step-by-step implementation guidance. Information localization hints, for instance, help find specific locations of edits by directly naming a file to be edited (``You would need to edit \texttt{test\_function()} in \texttt{src/testfile.py}''), help find necessary information (``Look at the README to find the descriptions of the hyperparameters''), or provide certain pieces of information that are part of the given input but nontrivial to find (``Use ID \#1014 for the special token''). On the other hand, the second level of hints breaks down the gold solution into concrete implementation steps. Therefore, we expect the second level of hints to yield substantially higher success rates. In our experiments, hints are cumulative; when providing the second level of hints, the first level of hints is also provided.

\subsection{Results}
\label{sec:results}
\paragraph{Main experiment} Figure~\ref{fig:main_results} shows our main results. Most agents struggle with the task, with the best performing agent (OpenHands + Claude 4 Sonnet) achieving a 33\% average final success rate. All agents achieved nonzero execution success rates except for DeepSeek-R1, which failed completely. Claude 4 Sonnet again performed best, with a execution rate of 68\% when combined with OpenHands. The agents overall achieved high file recall, showing that they were able to locate core edit targets based on the instructions.

\begin{figure*}
    \centering
\includegraphics[width=\linewidth]{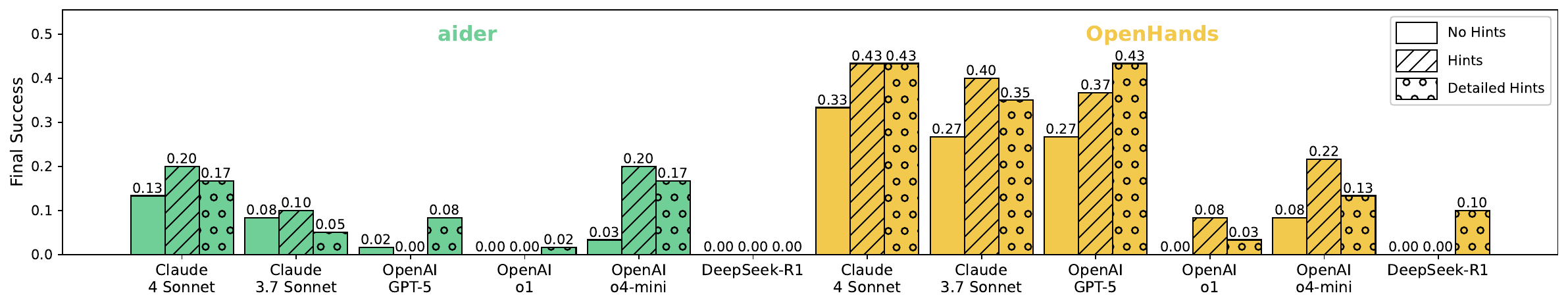}
    \caption{Final success rates for each agent-LLM combination and hint level.}
    \label{fig:hints-results}
\end{figure*}

\paragraph{Additional hints} \Cref{fig:hints-results} (and \Cref{tab:results-with-hintlevel} in \Cref{app:experiment-details}) shows the results of additional experiments with two different hint levels. Generally, hints improve the final success rate, but tend to help less when the default success rate was zero, suggesting there is a base level of competence required to make use of the hints provided. With the hints, we could boost the performance of two of the best agents, OpenHands + \{Claude 4 Sonnet, GPT-5\}, both achieving 43\% final success rates.
\subsection{Resource Consumption}
\begin{figure*}
    \centering
    \includegraphics[width=\linewidth]{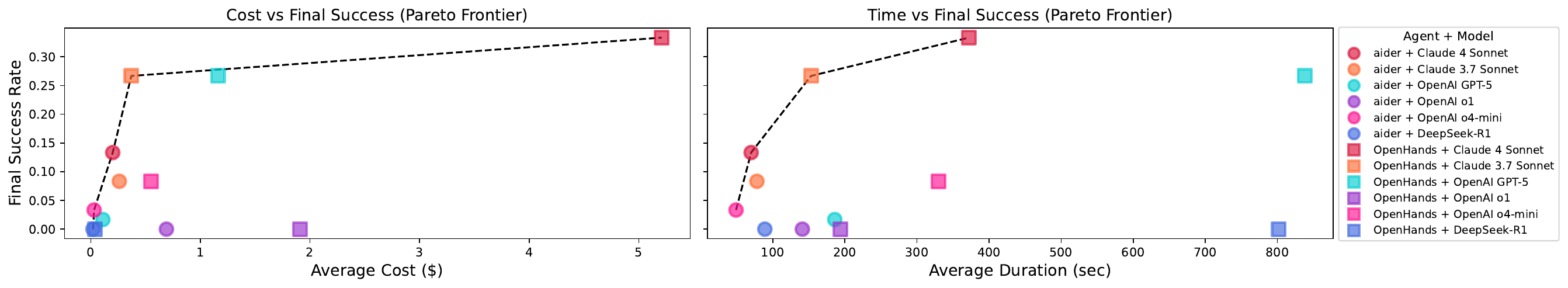}
    \caption{Cost effectiveness and time efficiency of coding agents on \benchmark{}.}
    \label{fig:pareto}
\end{figure*}

Based on the final success rate, we plot the cost/time vs. performance tradeoff (Figure~\ref{fig:pareto}), showing that aider + o4-mini, aider + Claude 4 Sonnet, OpenHands + Claude 3.7 Sonnet and OpenHands + Claude 4 Sonnet lie on the Pareto frontier for both cost and time. We provide the full time and cost estimates for agent runs in \Cref{app:experiment-details}, \Cref{tab:cost-duration}. In terms of token usage statistics, aider consistently used 2 turns due to its non-iterative design. OpenHands used more turns and therefore more tokens, especially with Claude 4 Sonnet, reaching up to 1.85M prompt tokens (almost 592 times more than aider). See Table~\ref{tab:token-usage} in \Cref{app:experiment-details} for token usage statistics by model and by hint levels. We also analyze the distribution of tool usage across backbone LLMs in \Cref{appendix:openhands_tool_usage}.
\section{Analysis and Discussion}
\subsection{Patterns of Error}
\label{sec:error-analysis}
We discuss notable error patterns, dividing them into explicit and implicit errors. We treat cases where the agent-generated code failed to execute as explicit errors, and cases where the execution succeeded but the experimental outcome did not match the numerical criteria as implicit. 

\paragraph{Explicit errors} 
Explicit errors were automatically identifiable from execution logs. The most common source of error was Python value errors (e.g., incorrect chat templates or invalid parameters). These errors were observed in all agents. Another common source of error was empty patch files due to the failure of the agent to modify any code. The majority of the empty patch file errors were from aider + \{DeepSeek-R1, o4-mini\}. We attribute this to the non-iterative nature of this agent framework: agents need to solve the entire extension task in one shot rather than breaking it down, often leading to incomplete or failed command executions during agent runs. Beyond these cases, most explicit errors were Python errors and they were mostly Python native errors rather than library-specific errors. Agents with Claude or GPT-5 as backbone led to fewer SyntaxErrors (in particular, OpenHands + \{Claude 4/3.7 Sonnet, GPT-5\} had no SyntaxErrors), whereas o1 produced SyntaxErrors frequently. There were also several cases of execution timeout, which occurs when the experiment runtime exceeds the limit of 12 hours we set (no gold solution required more than 6 hours). The full error distribution is shown in \Cref{fig:explicit_error_types} and Tables~\ref{tab:error_breakdown_1} and \ref{tab:error_breakdown_2} in Appendix~\ref{app:experiment-details}. 

\paragraph{Implicit errors} 
Analysis of implicit errors (execution success but mismatch with gold outcome) involved greater manual effort because it required a holistic review of agent edits. Therefore, we focused our analysis on the top 2 agents (OpenHands + \{Claude 4, GPT-5\}). Overall, the agents’ implicit errors were categorizable into errors in implementation logic and errors in value (e.g., within-bounds index errors, incorrect hyperparameters or paths)---the ratio of logic vs. value errors was about 2:1. We also estimated the debugging difficulty from the manually identified sources of error, using the scale of easy (requires small local fix), medium (requires logical but local revisions), and hard (requires holistic revisions). The majority of the errors were easy to debug (16 easy, 4 medium, 4 hard). Many of the medium and hard implicit errors arose from the agent ``over-editing'' the code beyond the given instructions (e.g., adding extra (incorrect) exception handling or changing irrelevant flags/prompts). These unrequested edits often caused silent failures or subtle deviations from the gold implementation leading to markedly different results, making debugging harder. We provide more detailed discussions in Appendix~\ref{app:qualitative-observations}.

\subsubsection{Qualitative Observations}

\paragraph{Implicit errors increase as model capacity increases, but are more difficult to analyze} A high-level observation is a general pitfall associated with stronger models (for our task and coding tasks more generally): the cause of failure is difficult to identify. Better models produced more implicit errors (e.g., OpenHands + Claude 3.7: 6, OpenHands + Claude 4: 24), where the code successfully executes but the outcome is incorrect. In such cases, the reasons behind failure were not always easily traceable even for the experts who implemented the solutions. This highlights the need for rigorous tests if an agent were to be deployed in practice. Plausible-looking implementations that execute can lead researchers to draw conclusions from faulty implementations, and over-reliance on coding agents may lead to a proliferation of incorrect results in the scientific literature.
\begin{figure}
    \centering
    \includegraphics[width=0.9\columnwidth]{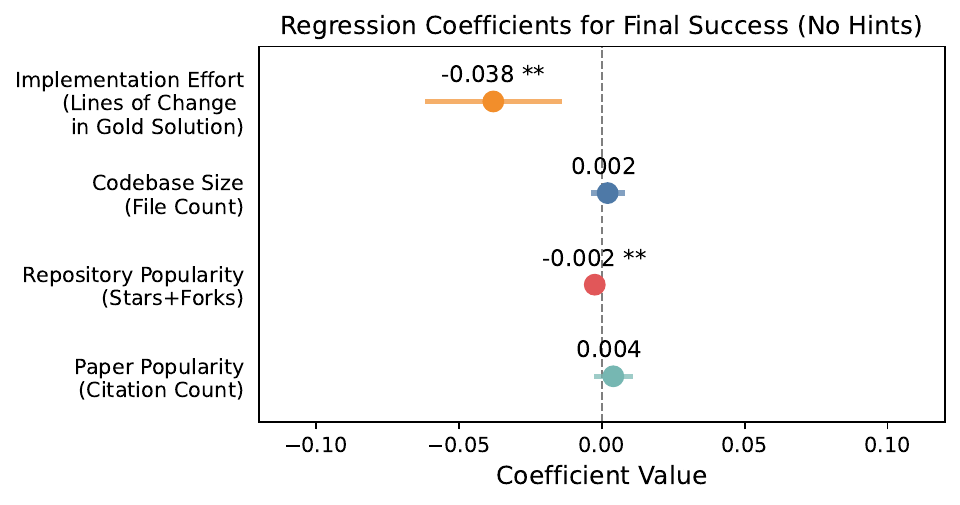}
    \caption{Regression coefficients with 95\% confidence intervals for predictors of final success. (Regression model: $\texttt{final\_success} \sim \texttt{line\_change} + \texttt{file\_count} + \texttt{repository\_popularity} + \texttt{citation} + (1 \mid \texttt{model}$)). ( $^{\ast\ast}$: $p<.01$)}
    \label{fig:coefficients-plot}
\end{figure}

\paragraph{Overthinking is often an issue} A prominent issue with weaker LLMs (Deepseek-R1, o1, o4-mini) was overthinking, where the thinking process was excessive both in terms of the number of output tokens and agent runtime, frequently leading to no actual output in terms of code generation. aider + DeepSeek-R1 was especially prone to this behavior, overthinking being one of the most prominent failure modes (close to one third of total failures). One possibility is that models' reasoning behavior somehow clashes with the reasoning/``thinking'' loop of the agent framework, but this pattern appears weaker in Claude 4/3.7 Sonnet and GPT-5, which also are reasoning models.

\paragraph{Agents vary in their ability to make use of hints} As noted in \Cref{sec:results}, providing additional hints did not always improve agents’ success rates, nor did more hints necessarily yield greater gains. While the best agent
(OpenHands + Claude 4) benefited from both levels of hints, many others showed no improvement. The best agents generally benefited more from hints, suggesting that a certain baseline competence or underlying model capacity may be required to leverage more detailed, human-written guidance. We observed idiosyncratic task-level variation as well; for instance, for the Othello task, OpenHands + Claude 3.7 Sonnet achieved 100\% success rate with no hints and with the first level of hints, but 0\% success rate when additionally given the second level of hints. Upon closer observation, the agent was unable to correctly use a helper function specified in the second level of hints, but with less detailed hints, it was able to correctly re-implement the helper function's logic. However, it was not the case that this particular hint was misleading, since the best performing agent was able to use this hint to achieve 100\% success rate on this task. This can be interpreted as models varying in their ability to implement different equally plausible solutions, and the step-by-step guideline in the second level of hints specifying a different solution from the one that the model could implement easily.

\subsection{What makes an extension difficult for agents?}
We hypothesize four sources of difficulty that could contribute to agent failure: (1) implementation effort; (2) codebase size; (3) unfamiliarity with the codebase; and (4) unfamiliarity with the research topic. We operationalize them as: (1) lines of code change in our gold solution; (2) file counts of the original codebase; (3) GitHub stars + forks (repository popularity); and (4) Google Scholar citations of the research paper(s), respectively. We use these as predictors of final success in a mixed-effects model with model identity as a random effect. Figure~\ref{fig:coefficients-plot} shows the regression coefficients. Lines of code changes has a significant negative effect ($\beta = -0.038$, $p < 0.01$) on final success, indicating that tasks with higher implementation effort are more difficult. Repository popularity had a significant effect but the effect size was negligible. Other factors were not statistically significant. We also compute cyclomatic complexity as a proxy for implementation difficulty but exclude it from the regression due to strong correlation with lines of code change (see Appendix~\ref{appendix:cyclomatic_complexity}).

\section{Conclusion}

We presented \benchmark{}, a benchmark evaluating the autonomous capacity of AI systems to implement hypothesis-driven research extensions in the domain of AI research. \benchmark{} consists of realistic but well-scoped extension tasks motivated by existing research. To perform well, a system must be able to understand the expert-written extension instructions situated in specific research context, understand the structure and logic of the original codebase, and autonomously plan and implement the requested extension. Our tasks are by design robust to data contamination due to the extensions requiring novel implementations whose solutions are not available publicly. Experiments with various agent frameworks combined with competent backbone LLMs show that most systems struggle on our benchmark, with the best performing model (OpenHands + Claude 4 Sonnet) achieving a 33\%  success rate. Notably, agents with o1 or DeepSeek-R1 as backbone showed (close to) zero success rate. Nevertheless, closer analysis of the best-performing agents revealed promise: agents using the strongest backbone LLMs (Claude 4 Sonnet, GPT-5) achieved higher execution success rates than those using weaker models, with implementations often syntactically valid and logically on the right track. 
Overall, this observation, taken together with the large headroom, highlights the utility of \benchmark{} for guiding future developments of research agents. We outline actionable modeling recommendations in \Cref{app:recommendations}. 

\paragraph{The future of \benchmark{}} Finally, as discussed in the introduction, we view the release of \benchmark{} as a start to a larger community-driven effort. While our tasks were primarily in the NLP domain, we believe the format of the extension task and the evaluation framework are broadly applicable. We hope the setup draws community interest in research extensions for evaluating agents and we welcome community contributions (\url{https://rexbench.com/}) for broader coverage of tasks.

\section*{Limitations}
 As discussed in \Cref{sec:task-definition}, a benchmark task being realistic inherently conflicts with the ease of automatic evaluation. In particular, a task like research extension can be extremely open ended in reality, even when constrained with a specific extension proposal and hypothesis. We opted for a middle ground where we do not enforce strong limitations on \textit{how} a system may implement the target extension and condition final success on alignment of numerical outcomes. This necessitated a stronger control for sources of variation, which led us to write instructions as self-contained and unambiguous as possible. This setting is idealized in that they are much more informative and clearer than an actual task a human researcher may face, even in scenarios where the extension idea is provided to them (e.g., an advisor suggesting to a PhD student ``How about we try X this time?''), missing out on the real difficulties lying in the initial trial-and-error concretization step. 
 
Furthermore, while the three automatic metrics that we provide 
measure different aspects of success, additional process-level metrics such as landmark evaluation \citep{xu2025theagentcompany,bogin2024super} would help alleviate the difficulty of post-hoc error analysis discussed in \Cref{sec:error-analysis}, especially for implicit errors, as well as reducing reward hacking or gamification of the benchmark. However, not all tasks are well-suited for such intermediate checks and additionally, their implementation requires substantial manual effort.

Additionally, the benchmark would benefit from a greater number of tasks. That being said, however, each task in our benchmark has high complexity as well as carrying signal; the current setup already clearly highlights performance differences between different agent setups and  different backbone models. While the benchmark may currently not be able to reliably estimate differences between models with highly similar levels of abilities, our results demonstrate that it is effective at tracking overall progress as new models are being released (e.g., Claude 3.7 to 4).

Regarding broader societal impacts, the baseline agents we developed for this work did not reach the level of competence that we believe would translate into autonomous research extension capacities in the real world. Still, our benchmark may contribute to developing such systems in the future, which may have positive impacts such as contributing to better replicability and faster iterations of empirical hypothesis verification. On the other hand, given the difficulty of debugging errors, deployment of such systems without rigorous verification measures faces the danger of leading researchers to draw conclusions from faulty implementations and of the erosion of trust in published results.

\section*{Ethical considerations}

In this work, we showed that current LLM-based agents cannot reliably produce code for novel AI research without additional human supervision. We based this argument on the low final success rate of all evaluated agents, as well as the danger of the increasing trend of implicit errors as model capacity improves. Given the rapid progress of AI research and model development, it is a likely possibility that new agents would perform significantly better on this benchmark in the near future. The biggest risk we therefore foresee is that good performance on this benchmark is seen as a sufficient condition for reliable agents rather than a necessary one. While we consider the benchmark to be well-suited for measuring progress in the development of future agents, good performance should NOT be seen as sufficient evidence for an agent being able to autonomously produce reliable research code.

Executing machine-written code always bears safety risks and providing AI agents with too much freedom for exploration may enable them to cause harm. To mitigate this risk, we narrowly scoped the implementation tasks in our experiments fully based on human-generated hypotheses and instructions. Furthermore, any machine-written code was executed in a containerized environment without internet access. We recommend similar setups for the execution of any code that is output by AI agents.

\section*{Acknowledgments}
This work was supported by funding from Good Ventures Foundation via Coefficient Giving awarded to NK and SS, from Google awarded to NK, and from WWTF through the project ``Understanding Language in Context'' (WWTF Vienna Research Group VRG23-007) awarded to SS. We acknowledge that the computational work reported on in this paper was performed on the Shared Computing Cluster which is administered by \href{https://www.bu.edu/tech/support/research/}{Boston University's Research Computing Services} and the shared computing cluster which is administered by \href{https://nerc.mghpcc.org}{New England Research Cloud (NERC)}. We additionally thank Augustine Abaris from BU SCC for technical advice, Max Nadeau and Ajeya Cotra from Coefficient Giving (then Open Philanthropy) for initial project advice, Dora Agali for contributing to agent experiments, and Zilu Tang for help with setting up the Explain then Translate task.

\bibliography{references}

@inproceedings{hua2025researchcodebench,
title={Research{C}ode{B}ench: Benchmarking {LLM}s on Implementing Novel Machine Learning Research Code},
author={Tianyu Hua and Harper Hua and Violet Xiang and Benjamin Klieger and Sang T. Truong and Weixin Liang and Fan-Yun Sun and Nick Haber},
booktitle={The Thirty-ninth Annual Conference on Neural Information Processing Systems Datasets and Benchmarks Track},
year={2025},
url={https://openreview.net/forum?id=3k70Vt0YFS}
}

@article{li2025deepcode,
  title={Deep{C}ode: Open Agentic Coding},
  author={Li, Zongwei and Li, Zhonghang and Guo, Zirui and Ren, Xubin and Huang, Chao},
  journal={arXiv:2512.07921},
  year={2025},
  url={https://arxiv.org/abs/2512.07921}
}

@inproceedings{nathani2025mlgym,
title={{MLG}ym: A New Framework and Benchmark for Advancing {AI} Research Agents},
author={Deepak Nathani and Lovish Madaan and Nicholas Roberts and Nikolay Bashlykov and Ajay Menon and Vincent Moens and Mikhail Plekhanov and Amar Budhiraja and Despoina Magka and Vladislav Vorotilov and Gaurav Chaurasia and Dieuwke Hupkes and Ricardo Silveira Cabral and Tatiana Shavrina and Jakob Nicolaus Foerster and Yoram Bachrach and William Yang Wang and Roberta Raileanu},
booktitle={Second Conference on Language Modeling},
year={2025},
url={https://openreview.net/forum?id=ryTr83DxRq}
}

@inproceedings{sharma2026researchrubrics,
title={Research{R}ubrics: A Benchmark of Prompts and Rubrics For Evaluating Deep Research Agents},
author={Manasi Sharma and Chen Bo Calvin Zhang and Chaithanya Bandi and Clinton Wang and Ankit Aich and Huy Nghiem and Tahseen Rabbani and Ye Htet and Brian Jang and Sumana Basu and Aishwarya Balwani and Denis Peskoff and Marcos Ayestaran and Sean M. Hendryx and Brad Kenstler and Bing Liu},
booktitle={The Fourteenth International Conference on Learning Representations},
year={2026},
url={https://openreview.net/forum?id=ErnvfmSX0P}
}

@article{miyai2026jr,
title={Jr. {AI} Scientist and Its Risk Report: Autonomous Scientific Exploration from a Baseline Paper},
author={Atsuyuki Miyai and Mashiro Toyooka and Takashi Otonari and Zaiying Zhao and Kiyoharu Aizawa},
journal={Transactions on Machine Learning Research},
issn={2835-8856},
year={2026},
url={https://openreview.net/forum?id=OeV062d8Sw},
note={}
}

@article{mitchener2025kosmos,
  title={Kosmos: An {AI} Scientist for Autonomous Discovery},
  author={Mitchener, Ludovico and Yiu, Angela and Chang, Benjamin and Bourdenx, Mathieu and Nadolski, Tyler and Sulovari, Arvis and Landsness, Eric C and Barabasi, Daniel L and Narayanan, Siddharth and Evans, Nicky and Shriya Reddy and Martha Foiani and Aizad Kamal and Leah P. Shriver and Fang Cao and Asmamaw T. Wassie and Jon M. Laurent and Edwin Melville-Green and Mayk Caldas and Albert Bou and Kaleigh F. Roberts and Sladjana Zagorac and Timothy C. Orr and Miranda E. Orr and Kevin J. Zwezdaryk and Ali E. Ghareeb and Laurie McCoy and Bruna Gomes and Euan A. Ashley and Karen E. Duff and Tonio Buonassisi and Tom Rainforth and Randall J. Bateman and Michael Skarlinski and Samuel G. Rodriques and Michaela M. Hinks and Andrew D. White},
  journal={arXiv:2511.02824},
  year={2025},
  url={https://arxiv.org/abs/2511.02824},
}

@article{yang2025codeclash,
  title={Code{C}lash: Benchmarking Goal-Oriented Software Engineering},
  author={Yang, John and Lieret, Kilian and Yang, Joyce and Jimenez, Carlos E and Press, Ofir and Schmidt, Ludwig and Yang, Diyi},
  journal={arXiv:2511.00839},
  year={2025},
  url={https://arxiv.org/abs/2511.00839},
}

@inproceedings{zhao2025abgen,
  title={{Ab{G}en}: Evaluating Large Language Models in Ablation Study Design and Evaluation for Scientific Research},
  author={Zhao, Yilun and Chen, Weiyuan and Xu, Zhijian and Patwardhan, Manasi and Wang, Chengye and Liu, Yixin and Vig, Lovekesh and Cohan, Arman},
  booktitle={Proceedings of the 63rd Annual Meeting of the Association for Computational Linguistics (Volume 1: Long Papers)},
  pages={12479--12491},
  year={2025},
  url={https://aclanthology.org/2025.acl-long.611/},
}

@inproceedings{miao2026recodeh,
title={{RECODE}-H: A Benchmark for Research Code Development with Interactive Human Feedback},
author={Chunyu Miao and Henry Peng Zou and Yangning Li and Yankai Chen and Yibo Wang and Fangxin Wang and Yifan Li and Wooseong Yang and Bowei He and Xinni Zhang and Dianzhi Yu and Hanchen Yang and Hoang H Nguyen and Yue Zhou and Jie Yang and Jizhou Guo and Wenzhe Fan and Chin-Yuan Yeh and Panpan Meng and Liancheng Fang and Jinhu Qi and Wei-Chieh Huang and Zhengyao Gu and Yuwei Han and Langzhou He and Yuyao Yang and Xue Liu and Irwin King and Philip S. Yu},
booktitle={The Fourteenth International Conference on Learning Representations},
year={2026},
url={https://openreview.net/forum?id=IKnuyyPHCV}
}

@article{liu2025guide,
  title={{GUIDE}: Towards Scalable Advising for Research Ideas},
  author={Liu, Yaowenqi and Meng, BingXu and Pan, Rui and Liu, Yuxing and Huang, Jerry and You, Jiaxuan and Zhang, Tong},
  journal={arXiv:2507.08870},
  year={2025},
  url={https://arxiv.org/abs/2507.08870},
}

@article{deng2025swe,
  title={{SWE}-{B}ench {P}ro: Can {AI} Agents Solve Long-Horizon Software Engineering Tasks?},
  author={Deng, Xiang and Da, Jeff and Pan, Edwin and He, Yannis Yiming and Ide, Charles and Garg, Kanak and Lauffer, Niklas and Park, Andrew and Pasari, Nitin and Rane, Chetan and Sampath, Karmini and Krishnan, Maya and Kundurthy, Srivatsa and  Hendryx, Sean and Wang, Zifan and Bharadwaj, Vijay and  Holm, Jeff and Aluri, Raja and Bo Calvin Zhang, Chen and  Jacobson, Noah and Liu, Bing and  Kenstler, Brad},
  journal={arXiv:2509.16941},
  year={2025},
  url={https://arxiv.org/abs/2509.16941}, 
}

@inproceedings{
    wang2024openhands,
    title={Open{H}ands: An Open Platform for {AI} Software Developers as Generalist Agents},
    author={Xingyao Wang and Boxuan Li and Yufan Song and Frank F. Xu and Xiangru Tang and Mingchen Zhuge and Jiayi Pan and Yueqi Song and Bowen Li and Jaskirat Singh and Hoang H. Tran and Fuqiang Li and Ren Ma and Mingzhang Zheng and Bill Qian and Yanjun Shao and Niklas Muennighoff and Yizhe Zhang and Binyuan Hui and Junyang Lin and Robert Brennan and Hao Peng and Heng Ji and Graham Neubig},
    booktitle={The Thirteenth International Conference on Learning Representations},
    year={2025},
    url={https://openreview.net/forum?id=OJd3ayDDoF}
}

@misc{aider2023github,
  author       = {aider AI},
  title        = {aider: {AI} pair programming in your terminal},
  year         = {2023},
  howpublished = {\url{https://github.com/Aider-AI/aider}},
  note         = {Accessed: 2025-05-12}
}

@inproceedings{
merrill2026terminalbench,
title={Terminal-{B}ench: Benchmarking Agents on Hard, Realistic Tasks in Command Line Interfaces},
author={Mike A Merrill and Alexander Glenn Shaw and Nicholas Carlini and Boxuan Li and Harsh Raj and Ivan Bercovich and Lin Shi and Jeong Yeon Shin and Thomas Walshe and E. Kelly Buchanan and Junhong Shen and Guanghao Ye and Haowei Lin and Jason Poulos and Maoyu Wang and Marianna Nezhurina and Di Lu and Orfeas Menis Mastromichalakis and Zhiwei Xu and Zizhao Chen and Yue Liu and Robert Zhang and Leon Liangyu Chen and Anurag Kashyap and Jan-Lucas Uslu and Jeffrey Li and Jianbo Wu and Minghao Yan and Song Bian and Vedang Sharma and Ke Sun and Steven Dillmann and Akshay Anand and Andrew Lanpouthakoun and Bardia Koopah and Changran Hu and Etash Kumar Guha and Gabriel H. S. Dreiman and Jiacheng Zhu and Karl Krauth and Li Zhong and Niklas Muennighoff and Robert Kwesi Amanfu and Shangyin Tan and Shreyas Pimpalgaonkar and Tushar Aggarwal and Xiangning Lin and Xin Lan and Xuandong Zhao and Yiqing Liang and Yuanli Wang and Zilong Wang and Changzhi Zhou and David Heineman and Hange Liu and Harsh Trivedi and John Yang and Junhong Lin and Manish Shetty and Michael Yang and Nabil Omi and Negin Raoof and Shanda Li and Terry Yue Zhuo and Wuwei Lin and Yiwei Dai and Yuxin Wang and Wenhao Chai and Shang Zhou and Dariush Wahdany and Ziyu She and Jiaming Hu and Zhikang Dong and Yuxuan Zhu and Sasha Cui and Ahson Saiyed and Arinbj{\"o}rn Kolbeinsson and Christopher Michael Rytting and Ryan Marten and Yixin Wang and Jenia Jitsev and Alex Dimakis and Andy Konwinski and Ludwig Schmidt},
booktitle={The Fourteenth International Conference on Learning Representations},
year={2026},
url={https://openreview.net/forum?id=a7Qa4CcHak}
}

@inproceedings{starace2025paperbench,
title={Paper{B}ench: Evaluating {AI}{\textquoteright}s Ability to Replicate {AI} Research},
author={Giulio Starace and Oliver Jaffe and Dane Sherburn and James Aung and Jun Shern Chan and Leon Maksin and Rachel Dias and Evan Mays and Benjamin Kinsella and Wyatt Thompson and Johannes Heidecke and Amelia Glaese and Tejal Patwardhan},
booktitle={Forty-second International Conference on Machine Learning},
year={2025},
url={https://openreview.net/forum?id=xF5PuTLPbn}
}

@inproceedings{li2025scilitllm,
    title={SciLit{LLM}: How to Adapt {LLM}s for Scientific Literature Understanding},
    author={Sihang Li and Jin Huang and Jiaxi Zhuang and Yaorui Shi and Xiaochen Cai and Mingjun Xu and Xiang Wang and Linfeng Zhang and Guolin Ke and Hengxing Cai},
    booktitle={The Thirteenth International Conference on Learning Representations},
    year={2025},
    url={https://openreview.net/forum?id=8dzKkeWUUb}
}

@inproceedings{tian2024scicode,
title={Sci{C}ode: A Research Coding Benchmark Curated by Scientists},
author={Minyang Tian and Luyu Gao and Dylan Zhang and Xinan Chen and Cunwei Fan and Xuefei Guo and Roland Haas and Pan Ji and Kittithat Krongchon and Yao Li and Shengyan Liu and Di Luo and Yutao Ma and HAO TONG and Kha Trinh and Chenyu Tian and Zihan Wang and Bohao Wu and Shengzhu Yin and Minhui Zhu and Kilian Lieret and Yanxin Lu and Genglin Liu and Yufeng Du and Tianhua Tao and Ofir Press and Jamie Callan and Eliu A Huerta and Hao Peng},
booktitle={The Thirty-eight Conference on Neural Information Processing Systems Datasets and Benchmarks Track},
year={2024},
url={https://openreview.net/forum?id=ADLaALtdoG}
}

@article{abramovich2025ablationbench,
  title={Ablation{B}ench: Evaluating Automated Planning of Ablations in Empirical {AI} Research},
  author={Abramovich, Talor and Chechik, Gal},
  journal={arXiv:2507.08038},
  year={2025},
  url={https://arxiv.org/abs/2507.08038},
}

@inproceedings{jansen-etal-2025-codescientist,
title = "{C}ode{S}cientist: End-to-End Semi-Automated Scientific Discovery with Code-based Experimentation",author = "Jansen, Peter and Tafjord, Oyvind and Radensky, Marissa and Siangliulue, Pao and Hope, Tom and Dalvi Mishra, Bhavana and Majumder, Bodhisattwa Prasad and Weld, Daniel S and Clark, Peter",editor = "Che, Wanxiang and Nabende, Joyce and Shutova, Ekaterina and Pilehvar, Mohammad Taher",booktitle = "Findings of the Association for Computational Linguistics: ACL 2025",month = jul,year = "2025",address = "Vienna, Austria",publisher = acl,url = anth # {2025.findings-acl.692/},doi = "10.18653/v1/2025.findings-acl.692",pages = "13370--13467",ISBN = "979-8-89176-256-5"}

@inproceedings{chen2025scienceagentbench,
title={Science{A}gent{B}ench: Toward Rigorous Assessment of Language Agents for Data-Driven Scientific Discovery},
author={Ziru Chen and Shijie Chen and Yuting Ning and Qianheng Zhang and Boshi Wang and Botao Yu and Yifei Li and Zeyi Liao and Chen Wei and Zitong Lu and Vishal Dey and Mingyi Xue and Frazier N. Baker and Benjamin Burns and Daniel Adu-Ampratwum and Xuhui Huang and Xia Ning and Song Gao and Yu Su and Huan Sun},
booktitle={The Thirteenth International Conference on Learning Representations},
year={2025},
url={https://openreview.net/forum?id=6z4YKr0GK6}
}

@article{skarlinski2024language,
  title={Language agents achieve superhuman synthesis of scientific knowledge},
  author={Skarlinski, Michael D. and Cox, Sam and Laurent, Jon M and Braza, James D. and Hinks, Michaela and Hammerling, Michael J. and Ponnapati, Manvitha and Rodriques, Samuel G. and White, Andrew D.},
  journal={arXiv:2409.13740},
  year={2024},
  url={https://arxiv.org/abs/2409.13740}
}

@inproceedings{badertdinov2025swerebench,
title={{SWE}-rebench: An Automated Pipeline for Task Collection and Decontaminated Evaluation of Software Engineering Agents},
author={Ibragim Badertdinov and Alexander Golubev and Maksim Nekrashevich and Anton Shevtsov and Simon Karasik and Andrei Andriushchenko and Maria Trofimova and Daria Litvintseva and Boris Yangel},
booktitle={The Thirty-ninth Annual Conference on Neural Information Processing Systems Datasets and Benchmarks Track},
year={2025},
url={https://openreview.net/forum?id=nMpJoVmRy1}
}

@inproceedings{gu2024blade,
    title = "{BLADE}: Benchmarking Language Model Agents for Data-Driven Science",
    author = "Gu, Ken  and
      Shang, Ruoxi  and
      Jiang, Ruien  and
      Kuang, Keying  and
      Lin, Richard-John  and
      Lyu, Donghe  and
      Mao, Yue  and
      Pan, Youran  and
      Wu, Teng  and
      Yu, Jiaqian  and
      Zhang, Yikun  and
      Zhang, Tianmai M.  and
      Zhu, Lanyi  and
      Merrill, Mike A  and
      Heer, Jeffrey  and
      Althoff, Tim",
    editor = "Al-Onaizan, Yaser  and
      Bansal, Mohit  and
      Chen, Yun-Nung",
    booktitle = "Findings of the Association for Computational Linguistics: EMNLP 2024",
    month = nov,
    year = "2024",
    address = "Miami, Florida, USA",
    publisher = "Association for Computational Linguistics",
    url = "https://aclanthology.org/2024.findings-emnlp.815/",
    doi = "10.18653/v1/2024.findings-emnlp.815",
    pages = "13936--13971"
}

@article{laurent2407lab,
  title={{LAB}-{B}ench: Measuring capabilities of language models for biology research},
  author={Laurent, Jon M. and Janizek, Joseph D. and Ruzo, Michael and Hinks, Michaela M. and Hammerling, Michael J. and Narayanan, Siddharth and Ponnapati, Manvitha and White, Andrew D. and Rodriques, Samuel G.},
  journal={arXiv:2407.10362},
  year={2024},
  url={https://arxiv.org/abs/2407.10362}
}

@article{ziems2023can,
    author = {Ziems, Caleb and Held, William and Shaikh, Omar and Chen, Jiaao and Zhang, Zhehao and Yang, Diyi},
    title = {Can Large Language Models Transform Computational Social
                    Science?},
    journal = {Computational Linguistics},
    volume = {50},
    number = {1},
    pages = {237-291},
    year = {2024},
    month = {03},
    issn = {0891-2017},
    doi = {10.1162/coli_a_00502},
    url = {https://doi.org/10.1162/coli\_a\_00502},
    eprint = {https://direct.mit.edu/coli/article-pdf/50/1/237/2367175/coli\_a\_00502.pdf},
}

@article{choi2024use,
  title={How to use large language models for empirical legal research},
  author={Choi, Jonathan H.},
  journal={Journal of Institutional and Theoretical Economics (JITE)},
  volume={180},
  number={2},
  pages={214--233},
  year={2024},
  publisher={Mohr Siebeck, T{\"u}bingen},
  url={https://ideas.repec.org/a/mhr/jinste/urndoi10.1628-jite-2024-0006.html}
}

@article{boiko2023autonomous,
  title={Autonomous chemical research with large language models},
  author={Boiko, Daniil A and MacKnight, Robert and Kline, Ben and Gomes, Gabe},
  journal={Nature},
  volume={624},
  number={7992},
  pages={570--578},
  year={2023},
  publisher={Nature Publishing Group UK London},
  url={https://www.nature.com/articles/s41586-023-06792-0}
}

@article{gottweis2025towards,
  title={Towards an {AI} co-scientist},
  author={Gottweis, Juraj and Weng, Wei-Hung and Daryin, Alexander and Tu, Tao and Palepu, Anil and Sirkovic, Petar and Myaskovsky, Artiom and Weissenberger, Felix and Rong, Keran and Tanno, Ryutaro and and Saab, Khaled and Popovici, Dan and Blum, Jacob and Zhang, Fan and Chou, Katherine and Hassidim, Avinatan and  Gokturk, Burak and  Vahdat, Amin and  Kohli, Pushmeet and Matias, Yossi and  Carroll, Andrew and  Kulkarni, Kavita and Tomasev, Nenad and  Guan, Yuan and  Dhillon, Vikram and Dhaval Vaishnav, Eeshit and  Lee, Byron and Costa, Tiago R D and  Penadés, José R and Peltz, Gary and Xu, Yunhan and Pawlosky, Annalisa and Karthikesalingam, Alan and Natarajan, Vivek},
  journal={arXiv:2502.18864},
  year={2025},
  url={https://arxiv.org/abs/2502.18864}
}

@article{kitano2021nobel,
  title={Nobel {T}uring {C}hallenge: creating the engine for scientific discovery},
  author={Kitano, Hiroaki},
  journal={NPJ systems biology and applications},
  volume={7},
  number={1},
  pages={29},
  year={2021},
  publisher={Nature Publishing Group UK London},
  url={https://www.nature.com/articles/s41540-021-00189-3}
}

@inproceedings{si2025can,
title={Can {LLM}s Generate Novel Research Ideas? A Large-Scale Human Study with 100+ {NLP} Researchers},
author={Chenglei Si and Diyi Yang and Tatsunori Hashimoto},
booktitle={The Thirteenth International Conference on Learning Representations},
year={2025},
url={https://openreview.net/forum?id=M23dTGWCZy}
}

@inproceedings{bogin2024super,
    title = "{SUPER}: Evaluating Agents on Setting Up and Executing Tasks from Research Repositories",
    author = "Bogin, Ben  and
      Yang, Kejuan  and
      Gupta, Shashank  and
      Richardson, Kyle  and
      Bransom, Erin  and
      Clark, Peter  and
      Sabharwal, Ashish  and
      Khot, Tushar",
    editor = "Al-Onaizan, Yaser  and
      Bansal, Mohit  and
      Chen, Yun-Nung",
    booktitle = "Proceedings of the 2024 Conference on Empirical Methods in Natural Language Processing",
    month = nov,
    year = "2024",
    address = "Miami, Florida, USA",
    publisher = "Association for Computational Linguistics",
    url = "https://aclanthology.org/2024.emnlp-main.702/",
    doi = "10.18653/v1/2024.emnlp-main.702",
    pages = "12622--12645"
}

@article{
siegel2024corebench,
    title={{CORE}-Bench: Fostering the Credibility of Published Research Through a Computational Reproducibility Agent Benchmark},
    author={Zachary S Siegel and Sayash Kapoor and Nitya Nadgir and Benedikt Stroebl and Arvind Narayanan},
    journal={Transactions on Machine Learning Research},
    issn={2835-8856},
    year={2024},
    url={https://openreview.net/forum?id=BsMMc4MEGS},
    note={}
}

@inproceedings{xiang2025scireplicatebench,
title={Sci{R}eplicate-{B}ench: Benchmarking {LLM}s in Agent-driven Algorithmic Reproduction from Research Papers},
author={Yanzheng Xiang and Hanqi Yan and Shuyin Ouyang and Lin Gui and Yulan He},
booktitle={Second Conference on Language Modeling},
year={2025},
url={https://openreview.net/forum?id=8LoPjpvWde}
}

@inproceedings{seo2026papercode,
title={Paper2{C}ode: Automating Code Generation from Scientific Papers in Machine Learning},
author={Minju Seo and Jinheon Baek and Seongyun Lee and Sung Ju Hwang},
booktitle={The Fourteenth International Conference on Learning Representations},
year={2026},
url={https://openreview.net/forum?id=3DcaUTjdKc}
}

@inproceedings{du2024llms,
    title = "{LLM}s Assist {NLP} Researchers: Critique Paper (Meta-)Reviewing",
    author = "Du, Jiangshu  and
      Wang, Yibo  and
      Zhao, Wenting  and
      Deng, Zhongfen  and
      Liu, Shuaiqi  and
      Lou, Renze  and
      Zou, Henry Peng  and
      Narayanan Venkit, Pranav  and
      Zhang, Nan  and
      Srinath, Mukund  and
      Zhang, Haoran Ranran  and
      Gupta, Vipul  and
      Li, Yinghui  and
      Li, Tao  and
      Wang, Fei  and
      Liu, Qin  and
      Liu, Tianlin  and
      Gao, Pengzhi  and
      Xia, Congying  and
      Xing, Chen  and
      Jiayang, Cheng  and
      Wang, Zhaowei  and
      Su, Ying  and
      Shah, Raj Sanjay  and
      Guo, Ruohao  and
      Gu, Jing  and
      Li, Haoran  and
      Wei, Kangda  and
      Wang, Zihao  and
      Cheng, Lu  and
      Ranathunga, Surangika  and
      Fang, Meng  and
      Fu, Jie  and
      Liu, Fei  and
      Huang, Ruihong  and
      Blanco, Eduardo  and
      Cao, Yixin  and
      Zhang, Rui  and
      Yu, Philip S.  and
      Yin, Wenpeng",
    editor = "Al-Onaizan, Yaser  and
      Bansal, Mohit  and
      Chen, Yun-Nung",
    booktitle = "Proceedings of the 2024 Conference on Empirical Methods in Natural Language Processing",
    month = nov,
    year = "2024",
    address = "Miami, Florida, USA",
    publisher = "Association for Computational Linguistics",
    url = "https://aclanthology.org/2024.emnlp-main.292/",
    doi = "10.18653/v1/2024.emnlp-main.292",
    pages = "5081--5099"
}

@article{lu2024ai,
  title={The {AI} Scientist: Towards Fully Automated Open-Ended Scientific Discovery},
  author={Lu, Chris and Lu, Cong and Tjarko Lange, Robert and Foerster, Jakob and Clune, Jeff and Ha, David},
  journal={arXiv:2408.06292},
  year={2024},
  url={https://arxiv.org/abs/2408.06292}
}

@article{kon2025curie,
  title={Curie: Toward Rigorous and Automated Scientific Experimentation with {AI} Agents},
  author={Kon, Patrick Tser Jern and Liu, Jiachen and Ding, Qiuyi and Qiu, Yiming and Yang, Zhenning and Huang, Yibo and Srinivasa, Jayanth and Lee, Myungjin and Chowdhury, Mosharaf and Chen, Ang},
  journal={arXiv:2502.16069},
  year={2025},
  url={https://arxiv.org/abs/2502.16069}
}

@inproceedings{chan2025mlebench,
title={{MLE}-bench: Evaluating Machine Learning Agents on Machine Learning Engineering},
author={Jun Shern Chan and Neil Chowdhury and Oliver Jaffe and James Aung and Dane Sherburn and Evan Mays and Giulio Starace and Kevin Liu and Leon Maksin and Tejal Patwardhan and Aleksander Madry and Lilian Weng},
booktitle={The Thirteenth International Conference on Learning Representations},
year={2025},
url={https://openreview.net/forum?id=6s5uXNWGIh}
}

@inproceedings{jing2025dsbench,
title={{DSB}ench: How Far Are Data Science Agents from Becoming Data Science Experts?},
author={Liqiang Jing and Zhehui Huang and Xiaoyang Wang and Wenlin Yao and Wenhao Yu and Kaixin Ma and Hongming Zhang and Xinya Du and Dong Yu},
booktitle={The Thirteenth International Conference on Learning Representations},
year={2025},
url={https://openreview.net/forum?id=DSsSPr0RZJ}
}

@inproceedings{huang2024mlagentbench,
  title={{MLA}gent{B}ench: evaluating language agents on machine learning experimentation},
  author={Huang, Qian and Vora, Jian and Liang, Percy and Leskovec, Jure},
  booktitle={Proceedings of the 41st International Conference on Machine Learning},
  pages={20271--20309},
  year={2024},
  url={https://proceedings.mlr.press/v235/huang24y.html}
}

@inproceedings{tian2024debugbench,
    title = "{D}ebug{B}ench: Evaluating Debugging Capability of Large Language Models",
    author = "Tian, Runchu  and
      Ye, Yining  and
      Qin, Yujia  and
      Cong, Xin  and
      Lin, Yankai  and
      Pan, Yinxu  and
      Wu, Yesai  and
      Haotian, Hui  and
      Weichuan, Liu  and
      Liu, Zhiyuan  and
      Sun, Maosong",
    editor = "Ku, Lun-Wei  and
      Martins, Andre  and
      Srikumar, Vivek",
    booktitle = "Findings of the Association for Computational Linguistics: ACL 2024",
    month = aug,
    year = "2024",
    address = "Bangkok, Thailand",
    publisher = "Association for Computational Linguistics",
    url = "https://aclanthology.org/2024.findings-acl.247/",
    doi = "10.18653/v1/2024.findings-acl.247",
    pages = "4173--4198"
}

@inproceedings{
    jimenezswe,
    title={{SWE}-bench: Can Language Models Resolve Real-world {G}it{H}ub Issues?},
    author={Carlos E. Jimenez and John Yang and Alexander Wettig and Shunyu Yao and Kexin Pei and Ofir Press and Karthik R. Narasimhan},
    booktitle={The Twelfth International Conference on Learning Representations},
    year={2024},
    url={https://openreview.net/forum?id=VTF8yNQM66}
}

@article{zhang2025datascibench,
  title={Data{S}ci{B}ench: An {LLM} agent benchmark for data science},
  author={Zhang, Dan and Zhoubian, Sining and Cai, Min and Li, Fengzu and Yang, Lekang and Wang, Wei and Dong, Tianjiao and Hu, Ziniu and Tang, Jie and Yue, Yisong},
  journal={arXiv:2502.13897},
  year={2025},
  url={https://arxiv.org/abs/2502.13897}
}

@article{gandhi2025boxinggym,
  title={Boxing{G}ym: Benchmarking Progress in Automated Experimental Design and Model Discovery},
  author={Gandhi, Kanishk and Li, Michael Y and Goodyear, Lyle and Li, Louise and Bhaskar, Aditi and Zaman, Mohammed and Goodman, Noah D},
  journal={arXiv:2501.01540},
  url={https://arxiv.org/abs/2501.01540},
  year={2025}
}

@article{padigela2025ml,
  title={{ML}-{D}ev-{B}ench: Comparative Analysis of {AI} Agents on {ML} development workflows},
  author={Padigela, Harshith and Shah, Chintan and Juyal, Dinkar},
  journal={arXiv:2502.00964},
  year={2025},
  url={https://arxiv.org/abs/2502.00964}
}

@inproceedings{lee-etal-2025-checkeval,
    title = "{C}heck{E}val: A reliable {LLM}-as-a-Judge framework for evaluating text generation using checklists",
    author = "Lee, Yukyung  and
      Kim, JoongHoon  and
      Kim, Jaehee  and
      Cho, Hyowon  and
      Kang, Jaewook  and
      Kang, Pilsung  and
      Kim, Najoung",
    editor = "Christodoulopoulos, Christos  and
      Chakraborty, Tanmoy  and
      Rose, Carolyn  and
      Peng, Violet",
    booktitle = "Proceedings of the 2025 Conference on Empirical Methods in Natural Language Processing",
    month = nov,
    year = "2025",
    address = "Suzhou, China",
    publisher = "Association for Computational Linguistics",
    url = "https://aclanthology.org/2025.emnlp-main.796/",
    doi = "10.18653/v1/2025.emnlp-main.796",
    pages = "15771--15798",
    ISBN = "979-8-89176-332-6"
}

@inproceedings{kim2020cogs,
    title = "{COGS}: A Compositional Generalization Challenge Based on Semantic Interpretation",
    author = "Kim, Najoung  and
      Linzen, Tal",
    editor = "Webber, Bonnie  and
      Cohn, Trevor  and
      He, Yulan  and
      Liu, Yang",
    booktitle = "Proceedings of the 2020 Conference on Empirical Methods in Natural Language Processing (EMNLP)",
    month = nov,
    year = "2020",
    address = "Online",
    publisher = "Association for Computational Linguistics",
    url = "https://aclanthology.org/2020.emnlp-main.731/",
    doi = "10.18653/v1/2020.emnlp-main.731",
    pages = "9087--9105"
}

@inproceedings{csordas2021devil,
    title = "The Devil is in the Detail: Simple Tricks Improve Systematic Generalization of Transformers",
    author = "Csord{\'a}s, R{\'o}bert  and
      Irie, Kazuki  and
      Schmidhuber, Juergen",
    editor = "Moens, Marie-Francine  and
      Huang, Xuanjing  and
      Specia, Lucia  and
      Yih, Scott Wen-tau",
    booktitle = "Proceedings of the 2021 Conference on Empirical Methods in Natural Language Processing",
    month = nov,
    year = "2021",
    address = "Online and Punta Cana, Dominican Republic",
    publisher = "Association for Computational Linguistics",
    url = "https://aclanthology.org/2021.emnlp-main.49/",
    doi = "10.18653/v1/2021.emnlp-main.49",
    pages = "619--634"
}

@article{kim2024code,
  title={Code pretraining improves entity tracking abilities of language models},
  author={Kim, Najoung and Schuster, Sebastian and Toshniwal, Shubham},
  journal={arXiv:2405.21068},
  year={2024},
  url={https://arxiv.org/abs/2405.21068}
}

@inproceedings{tang2023explain,
    title = "Explain-then-translate: an analysis on improving program translation with self-generated explanations",
    author = "Tang, Zilu  and
      Agarwal, Mayank  and
      Shypula, Alexander  and
      Wang, Bailin  and
      Wijaya, Derry  and
      Chen, Jie  and
      Kim, Yoon",
    editor = "Bouamor, Houda  and
      Pino, Juan  and
      Bali, Kalika",
    booktitle = "Findings of the Association for Computational Linguistics: EMNLP 2023",
    month = dec,
    year = "2023",
    address = "Singapore",
    publisher = "Association for Computational Linguistics",
    url = "https://aclanthology.org/2023.findings-emnlp.119/",
    doi = "10.18653/v1/2023.findings-emnlp.119",
    pages = "1741--1788"
}

@article{hewitt2024instruction,
  title={Instruction following without instruction tuning},
  author={Hewitt, John and Liu, Nelson F. and Liang, Percy and Manning, Christopher D.},
  journal={arXiv:2409.14254},
  year={2024},
  url={https://arxiv.org/abs/2409.14254}
}

@inproceedings{kallini2024mission,
    title = "Mission: Impossible Language Models",
    author = "Kallini, Julie  and
      Papadimitriou, Isabel  and
      Futrell, Richard  and
      Mahowald, Kyle  and
      Potts, Christopher",
    editor = "Ku, Lun-Wei  and
      Martins, Andre  and
      Srikumar, Vivek",
    booktitle = "Proceedings of the 62nd Annual Meeting of the Association for Computational Linguistics (Volume 1: Long Papers)",
    month = aug,
    year = "2024",
    address = "Bangkok, Thailand",
    publisher = "Association for Computational Linguistics",
    url = "https://aclanthology.org/2024.acl-long.787/",
    doi = "10.18653/v1/2024.acl-long.787",
    pages = "14691--14714"
}

@inproceedings{
    li2023emergent,
    title={Emergent World Representations: Exploring a Sequence Model Trained on a Synthetic Task},
    author={Kenneth Li and Aspen K. Hopkins and David Bau and Fernanda Vi{\'e}gas and Hanspeter Pfister and Martin Wattenberg},
    booktitle={The Eleventh International Conference on Learning Representations},
    year={2023},
    url={https://openreview.net/forum?id=DeG07_TcZvT}
}

@inproceedings{nanda2023emergent,
    title = "Emergent Linear Representations in World Models of Self-Supervised Sequence Models",
    author = "Nanda, Neel  and
      Lee, Andrew  and
      Wattenberg, Martin",
    editor = "Belinkov, Yonatan  and
      Hao, Sophie  and
      Jumelet, Jaap  and
      Kim, Najoung  and
      McCarthy, Arya  and
      Mohebbi, Hosein",
    booktitle = "Proceedings of the 6th BlackboxNLP Workshop: Analyzing and Interpreting Neural Networks for NLP",
    month = dec,
    year = "2023",
    address = "Singapore",
    publisher = "Association for Computational Linguistics",
    url = "https://aclanthology.org/2023.blackboxnlp-1.2/",
    doi = "10.18653/v1/2023.blackboxnlp-1.2",
    pages = "16--30"
}

@inproceedings{wu2024reasoning,
    title = "Reasoning or Reciting? Exploring the Capabilities and Limitations of Language Models Through Counterfactual Tasks",
    author = {Wu, Zhaofeng  and
      Qiu, Linlu  and
      Ross, Alexis  and
      Aky{\"u}rek, Ekin  and
      Chen, Boyuan  and
      Wang, Bailin  and
      Kim, Najoung  and
      Andreas, Jacob  and
      Kim, Yoon},
    editor = "Duh, Kevin  and
      Gomez, Helena  and
      Bethard, Steven",
    booktitle = "Proceedings of the 2024 Conference of the North American Chapter of the Association for Computational Linguistics: Human Language Technologies (Volume 1: Long Papers)",
    month = jun,
    year = "2024",
    address = "Mexico City, Mexico",
    publisher = "Association for Computational Linguistics",
    url = "https://aclanthology.org/2024.naacl-long.102/",
    doi = "10.18653/v1/2024.naacl-long.102",
    pages = "1819--1862"
}

@inproceedings{xu2024re,
    title = "Re-Reading Improves Reasoning in Large Language Models",
    author = "Xu, Xiaohan  and
      Tao, Chongyang  and
      Shen, Tao  and
      Xu, Can  and
      Xu, Hongbo  and
      Long, Guodong  and
      Lou, Jian-Guang  and
      Ma, Shuai",
    editor = "Al-Onaizan, Yaser  and
      Bansal, Mohit  and
      Chen, Yun-Nung",
    booktitle = "Proceedings of the 2024 Conference on Empirical Methods in Natural Language Processing",
    month = nov,
    year = "2024",
    address = "Miami, Florida, USA",
    publisher = "Association for Computational Linguistics",
    url = "https://aclanthology.org/2024.emnlp-main.871/",
    doi = "10.18653/v1/2024.emnlp-main.871",
    pages = "15549--15575"
}

@inproceedings{yao2023tree,
 author = {Yao, Shunyu and Yu, Dian and Zhao, Jeffrey and Shafran, Izhak and Griffiths, Tom and Cao, Yuan and Narasimhan, Karthik},
 booktitle = {Advances in Neural Information Processing Systems},
 editor = {A. Oh and T. Naumann and A. Globerson and K. Saenko and M. Hardt and S. Levine},
 pages = {11809--11822},
 publisher = {Curran Associates, Inc.},
 title = {Tree of Thoughts: Deliberate Problem Solving with Large Language Models},
 url = {https://proceedings.neurips.cc/paper_files/paper/2023/file/271db9922b8d1f4dd7aaef84ed5ac703-Paper-Conference.pdf},
 volume = {36},
 year = {2023}
}

@inproceedings{weber2024varierr,
    title = "{V}ari{E}rr {NLI}: Separating Annotation Error from Human Label Variation",
    author = "Weber-Genzel, Leon  and
      Peng, Siyao  and
      De Marneffe, Marie-Catherine  and
      Plank, Barbara",
    editor = "Ku, Lun-Wei  and
      Martins, Andre  and
      Srikumar, Vivek",
    booktitle = "Proceedings of the 62nd Annual Meeting of the Association for Computational Linguistics (Volume 1: Long Papers)",
    month = aug,
    year = "2024",
    address = "Bangkok, Thailand",
    publisher = "Association for Computational Linguistics",
    url = "https://aclanthology.org/2024.acl-long.123/",
    doi = "10.18653/v1/2024.acl-long.123",
    pages = "2256--2269"
}

@inproceedings{eisenschlos2022winodict,
    title = "{W}ino{D}ict: Probing language models for in-context word acquisition",
    author = "Eisenschlos, Julian Martin  and
      Cole, Jeremy R.  and
      Liu, Fangyu  and
      Cohen, William W.",
    editor = "Vlachos, Andreas  and
      Augenstein, Isabelle",
    booktitle = "Proceedings of the 17th Conference of the European Chapter of the Association for Computational Linguistics",
    month = may,
    year = "2023",
    address = "Dubrovnik, Croatia",
    publisher = "Association for Computational Linguistics",
    url = "https://aclanthology.org/2023.eacl-main.7/",
    doi = "10.18653/v1/2023.eacl-main.7",
    pages = "94--102"
}

@article{luo2025llm4sr,
  title={{LLM4SR}: A Survey on Large Language Models for Scientific Research},
  author={Luo, Ziming and Yang, Zonglin and Xu, Zexin and Yang, Wei and Du, Xinya},
  journal={arXiv:2501.04306},
  year={2025},
  url={https://arxiv.org/abs/2501.04306}
}

@article{guo2025deepseek,
  title={Deep{S}eek-{R1}: Incentivizing reasoning capability in {LLM}s via reinforcement learning},
  author={Guo, Daya and Yang, Dejian and Zhang, Haowei and Song, Junxiao and Wang, Peiyi and Zhu, Qihao and Xu, Runxin and Zhang, Ruoyu and Ma, Shirong and Bi, Xiao and Zhang, Xiaokang and Yu, Xingkai and Wu, Yu and Wu, Z. F. and Gou, Zhibin and Shao, Zhihong and Li, Zhuoshu and Gao, Ziyi and Liu, Aixin and Xue, Bing and Wang, Bingxuan and Wu, Bochao and Feng, Bei and Lu, Chengda and Zhao, Chenggang and Deng, Chengqi and Ruan, Chong and Dai, Damai and Chen, Deli and Ji, Dongjie and Li, Erhang and Lin, Fangyun and Dai, Fucong and Luo, Fuli and Hao, Guangbo and Chen, Guanting and Li, Guowei and Zhang, H. and Xu, Hanwei and Ding, Honghui and Gao, Huazuo and Qu, Hui and Li, Hui and Guo, Jianzhong and Li, Jiashi and Chen, Jingchang and Yuan, Jingyang and Tu, Jinhao and Qiu, Junjie and Li, Junlong and Cai, J. L. and Ni, Jiaqi and Liang, Jian and Chen, Jin and Dong, Kai and Hu, Kai and You, Kaichao and Gao, Kaige and Guan, Kang and Huang, Kexin and Yu, Kuai and Wang, Lean and Zhang, Lecong and Zhao, Liang and Wang, Litong and Zhang, Liyue and Xu, Lei and Xia, Leyi and Zhang, Mingchuan and Zhang, Minghua and Tang, Minghui and Zhou, Mingxu and Li, Meng and Wang, Miaojun and Li, Mingming and Tian, Ning and Huang, Panpan and Zhang, Peng and Wang, Qiancheng and Chen, Qinyu and Du, Qiushi and Ge, Ruiqi and Zhang, Ruisong and Pan, Ruizhe and Wang, Runji and Chen, R. J. and Jin, R. L. and Chen, Ruyi and Lu, Shanghao and Zhou, Shangyan and Chen, Shanhuang and Ye, Shengfeng and Wang, Shiyu and Yu, Shuiping and Zhou, Shunfeng and Pan, Shuting and Li, S. S. and Zhou, Shuang and Wu, Shaoqing and Yun, Tao and Pei, Tian and Sun, Tianyu and Wang, T. and Zeng, Wangding and Liu, Wen and Liang, Wenfeng and Gao, Wenjun and Yu, Wenqin and Zhang, Wentao and Xiao, W. L. and An, Wei and Liu, Xiaodong and Wang, Xiaohan and Chen, Xiaokang and Nie, Xiaotao and Cheng, Xin and Liu, Xin and Xie, Xin and Liu, Xingchao and Yang, Xinyu and Li, Xinyuan and Su, Xuecheng and Lin, Xuheng and Li, X. Q. and Jin, Xiangyue and Shen, Xiaojin and Chen, Xiaosha and Sun, Xiaowen and Wang, Xiaoxiang and Song, Xinnan and Zhou, Xinyi and Wang, Xianzu and Shan, Xinxia and Li, Y. K. and Wang, Y. Q. and Wei, Y. X. and Zhang, Yang and Xu, Yanhong and Li, Yao and Zhao, Yao and Sun, Yaofeng and Wang, Yaohui and Yu, Yi and Zhang, Yichao and Shi, Yifan and Xiong, Yiliang and He, Ying and Piao, Yishi and Wang, Yisong and Tan, Yixuan and Ma, Yiyang and Liu, Yiyuan and Guo, Yongqiang and Ou, Yuan and Wang, Yuduan and Gong, Yue and Zou, Yuheng and He, Yujia and Xiong, Yunfan and Luo, Yuxiang and You, Yuxiang and Liu, Yuxuan and Zhou, Yuyang and Zhu, Y. X. and Huang, Yanping and Li, Yaohui and Zheng, Yi and Zhu, Yuchen and Ma, Yunxian and Tang, Ying and Zha, Yukun and Yan, Yuting and Ren, Z. Z. and Ren, Zehui and Sha, Zhangli and Fu, Zhe and Xu, Zhean and Xie, Zhenda and Zhang, Zhengyan and Hao, Zhewen and Ma, Zhicheng and Yan, Zhigang and Wu, Zhiyu and Gu, Zihui and Zhu, Zijia and Liu, Zijun and Li, Zilin and Xie, Ziwei and Song, Ziyang and Pan, Zizheng and Huang, Zhen and Xu, Zhipeng and Zhang, Zhongyu and Zhang, Zhen},
  journal={arXiv:2501.12948},
  year={2025},
  url={https://arxiv.org/abs/2501.12948}
}

@article{jaech2024openai,
  title={Open{AI} o1 system card},
  author={Aaron Jaech and Adam Kalai and Adam Lerer and Adam Richardson and Ahmed El-Kishky and Aiden Low and Alec Helyar and Aleksander Madry and Alex Beutel and Alex Carney and Alex Iftimie and Alex Karpenko and Alex Tachard Passos and Alexander Neitz and Alexander Prokofiev and Alexander Wei and Allison Tam and Ally Bennett and Ananya Kumar and Andre Saraiva and Andrea Vallone and Andrew Duberstein and Andrew Kondrich and Andrey Mishchenko and Andy Applebaum and Angela Jiang and Ashvin Nair and Barret Zoph and Behrooz Ghorbani and Ben Rossen and Benjamin Sokolowsky and Boaz Barak and Bob McGrew and Borys Minaiev and Botao Hao and Bowen Baker and Brandon Houghton and Brandon McKinzie and Brydon Eastman and Camillo Lugaresi and Cary Bassin and Cary Hudson and Chak Ming Li and Charles de Bourcy and Chelsea Voss and Chen Shen and Chong Zhang and Chris Koch and Chris Orsinger and Christopher Hesse and Claudia Fischer and Clive Chan and Dan Roberts and Daniel Kappler and Daniel Levy and Daniel Selsam and David Dohan and David Farhi and David Mely and David Robinson and Dimitris Tsipras and Doug Li and Dragos Oprica and Eben Freeman and Eddie Zhang and Edmund Wong and Elizabeth Proehl and Enoch Cheung and Eric Mitchell and Eric Wallace and Erik Ritter and Evan Mays and Fan Wang and Felipe Petroski Such and Filippo Raso and Florencia Leoni and Foivos Tsimpourlas and Francis Song and Fred von Lohmann and Freddie Sulit and Geoff Salmon and Giambattista Parascandolo and Gildas Chabot and Grace Zhao and Greg Brockman and Guillaume Leclerc and Hadi Salman and Haiming Bao and Hao Sheng and Hart Andrin and Hessam Bagherinezhad and Hongyu Ren and Hunter Lightman and Hyung Won Chung and Ian Kivlichan and Ian O'Connell and Ian Osband and Ignasi Clavera Gilaberte and Ilge Akkaya and Ilya Kostrikov and Ilya Sutskever and Irina Kofman and Jakub Pachocki and James Lennon and Jason Wei and Jean Harb and Jerry Twore and Jiacheng Feng and Jiahui Yu and Jiayi Weng and Jie Tang and Jieqi Yu and Joaquin Quiñonero Candela and Joe Palermo and Joel Parish and Johannes Heidecke and John Hallman and John Rizzo and Jonathan Gordon and Jonathan Uesato and Jonathan Ward and Joost Huizinga and Julie Wang and Kai Chen and Kai Xiao and Karan Singhal and Karina Nguyen and Karl Cobbe and Katy Shi and Kayla Wood and Kendra Rimbach and Keren Gu-Lemberg and Kevin Liu and Kevin Lu and Kevin Stone and Kevin Yu and Lama Ahmad and Lauren Yang and Leo Liu and Leon Maksin and Leyton Ho and Liam Fedus and Lilian Weng and Linden Li and Lindsay McCallum and Lindsey Held and Lorenz Kuhn and Lukas Kondraciuk and Lukasz Kaiser and Luke Metz and Madelaine Boyd and Maja Trebacz and Manas Joglekar and Mark Chen and Marko Tintor and Mason Meyer and Matt Jones and Matt Kaufer and Max Schwarzer and Meghan Shah and Mehmet Yatbaz and Melody Y. Guan and Mengyuan Xu and Mengyuan Yan and Mia Glaese and Mianna Chen and Michael Lampe and Michael Malek and Michele Wang and Michelle Fradin and Mike McClay and Mikhail Pavlov and Miles Wang and Mingxuan Wang and Mira Murati and Mo Bavarian and Mostafa Rohaninejad and Nat McAleese and Neil Chowdhury and Neil Chowdhury and Nick Ryder and Nikolas Tezak and Noam Brown and Ofir Nachum and Oleg Boiko and Oleg Murk and Olivia Watkins and Patrick Chao and Paul Ashbourne and Pavel Izmailov and Peter Zhokhov and Rachel Dias and Rahul Arora and Randall Lin and Rapha Gontijo Lopes and Raz Gaon and Reah Miyara and Reimar Leike and Renny Hwang and Rhythm Garg and Robin Brown and Roshan James and Rui Shu and Ryan Cheu and Ryan Greene and Saachi Jain and Sam Altman and Sam Toizer and Sam Toyer and Samuel Miserendino and Sandhini Agarwal and Santiago Hernandez and Sasha Baker and Scott McKinney and Scottie Yan and Shengjia Zhao and Shengli Hu and Shibani Santurkar and Shraman Ray Chaudhuri and Shuyuan Zhang and Siyuan Fu and Spencer Papay and Steph Lin and Suchir Balaji and Suvansh Sanjeev and Szymon Sidor and Tal Broda and Aidan Clark and Tao Wang and Taylor Gordon and Ted Sanders and Tejal Patwardhan and Thibault Sottiaux and Thomas Degry and Thomas Dimson and Tianhao Zheng and Timur Garipov and Tom Stasi and Trapit Bansal and Trevor Creech and Troy Peterson and Tyna Eloundou and Valerie Qi and Vineet Kosaraju and Vinnie Monaco and Vitchyr Pong and Vlad Fomenko and Weiyi Zheng and Wenda Zhou and Wes McCabe and Wojciech Zaremba and Yann Dubois and Yinghai Lu and Yining Chen and Young Cha and Yu Bai and Yuchen He and Yuchen Zhang and Yunyun Wang and Zheng Shao and Zhuohan Li},
  journal={arXiv:2412.16720},
  year={2024},
  url={https://arxiv.org/abs/2412.16720}
}

@misc{openai2025gpt5,
  author       = {OpenAI},
  title        = {{GPT}‑5 System Card},
  year         = {2025},
  howpublished = {\url{https://cdn.openai.com/gpt-5-system-card.pdf}},
  note         = {Accessed: 2025-09-22}
}

@misc{anthropic2024claude37sonnet,
  author       = {Anthropic},
  title        = {{C}laude 3.7 {S}onnet System Card},
  year         = {2024},
  howpublished = {\url{https://assets.anthropic.com/m/785e231869ea8b3b/original/claude-3-7-sonnet-system-card.pdf}},
  note         = {Accessed: 2025-05-14}
}

@misc{anthropic2025claude4sonnet,
  author       = {Anthropic},
  title        = {{C}laude 4 {S}onnet System Card},
  year         = {2025},
  howpublished = {\url{https://www-cdn.anthropic.com/6d8a8055020700718b0c49369f60816ba2a7c285.pdf}},
  note         = {Accessed: 2025-09-22}
}

@inproceedings{xu2025theagentcompany,
title={The{A}gent{C}ompany: Benchmarking {LLM} Agents on Consequential Real World Tasks},
author={Frank F. Xu and Yufan Song and Boxuan Li and Yuxuan Tang and Kritanjali Jain and Mengxue Bao and Zora Zhiruo Wang and Xuhui Zhou and Zhitong Guo and Murong Cao and Mingyang Yang and Hao Yang Lu and Amaad Martin and Zhe Su and Leander Melroy Maben and Raj Mehta and Wayne Chi and Lawrence Keunho Jang and Yiqing Xie and Shuyan Zhou and Graham Neubig},
booktitle={The Thirty-ninth Annual Conference on Neural Information Processing Systems Datasets and Benchmarks Track},
year={2025},
url={https://openreview.net/forum?id=LZnKNApvhG}
}

@misc{singularity_2021,
  author       = {Singularity},
  title        = {Singularity},
  year         = 2021,
  publisher    = {Zenodo},
  doi          = {10.5281/zenodo.1310023},
  url          = {https://doi.org/10.5281/zenodo.1310023}
}

\appendix
\newpage
\section{Actionable recommendations}
\label{app:recommendations}

Based on the analyses of the agents tested in this work, we put forward several actionable recommendations for the future.
\paragraph{Short-horizon recommendations:}
\begin{itemize}[itemsep=0pt]
    \vspace{-1ex}
    \item \textbf{Incorporate iterative design}: Our findings show that iterative design is critical for success on our tasks: aider (a non-iterative framework) showed weaker performance in general, and many success scenarios for multi-turn agents could be attributed to effective use of the previous turns' output. For instance, in the CheckEval task, OpenHands + GPT-5 used one turn to inspect a file's structure with bash before writing code in the next.
    \item \textbf{Support scratchpads}: Agents frequently failed on the basis of small errors such as path mis-specification. Such errors could be easily caught if agents can make use of a ``scratchpad'' where small code snippets can be executed. 
    \item \textbf{Support ``repair'' mechanism}: Agents should incorporate a mechanism to repair a step in their action trajectory, for instance by reverting the changes made in the step and re-initiating the LLM call. One use case of this would be detecting and repairing overthinking in the LLM output, which was a prominent failure mode in several agents, especially with DeepSeek-R1 as backbone, that resulted in no code edits.
\end{itemize}

\paragraph{Longer-horizon recommendations:}
\begin{itemize}[itemsep=0pt]
    \vspace{-1ex}
    \item \textbf{More stringent verification}: One of the most concerning observations from our analysis is the increasing trend of implicit errors as the capacity of the backbone LLM grows. Under a benchmarking setup, numeric mismatches of the outcome to the gold solution easily indicates failure, but in real deployment scenarios, there exist no gold solutions. This indicates a need for more stringent verification processes, ideally by agent design rather than relying on manual verification from the end users.
    \item \textbf{Prevent over-editing}: A prominent failure mode of the strongest agents was ``over-editing'', where agents make unrequested modifications that often lead to implicit errors. Our findings show that simply instructing an agent to ``keep everything else not specified constant'' (see Appendix~\ref{appendix:winodict_instructions}) is insufficient. A general improvement in hallucination reduction and instruction-following would help, but for research coding where fine-grained controls of experimental details is critical, a more targeted solution for over-editing may be beneficial. 
    \item \textbf{Improve handling of long contexts}: Our analysis shows that the most important factor to agent failure is the size of the required edits. Given that the maximum lines of change in the gold solutions in our benchmark is not huge (in the magnitude of hundreds), there is a need for future agents to handle long contexts better, both within and across file boundaries.
\end{itemize}
\vspace{-2ex}

\section{Detailed Agent Configurations}
\label{app:agent-configurations}

\begin{table}[h]
    \centering
    \footnotesize
    \renewcommand{\arraystretch}{1.1}
    \setlength{\tabcolsep}{10pt}
    \resizebox{0.85\columnwidth}{!}{%
        \begin{tabular}{lcc} 
        \toprule
        Component         & aider  & OpenHands \\
        \midrule
        Repo navigation   & $\times$                & $\checkmark$         \\
        Tool use          & $\times$                & $\checkmark$         \\
        Bash execution    & $\times$                & $\checkmark$         \\
        Python execution  & $\times$                & $\times$         \\
        \bottomrule
        \end{tabular}}
    \caption{Agent components}
    \label{tab:agent-components}
\end{table}
\Cref{tab:agent-components} provides an overview of what kind of abilities each agent has.

\paragraph{aider} \href{https://github.com/Aider-AI/aider}{aider} is an open-source agent framework. We implemented our most basic agent based on aider, using the ``diff'' edit format where the LLM specifies file changes as search/replace blocks. We allowed up to 5 retries to handle API-side overload errors. Since aider lacks built-in file search capabilities, we added a preliminary stage where the LLM is given the codebase's directory tree along with the task instruction to identify files requiring modification. Unlike OpenHands, our aider implementation does not use bash execution or tools.

\paragraph{OpenHands} \href{https://github.com/All-Hands-AI/OpenHands/}{OpenHands} is an open-source agent framework that uses an LLM to control a range of pre-defined tools for understanding and modifying codebases. We modified the system prompt and the agent to disable execution of Python code. The agent was allowed to execute bash commands such as \texttt{grep} and \texttt{cat}, browse the web, load PDFs in a browser (if compatible with the backbone LLM), and edit files. We prompted this agent with the same one-line instruction.
We evaluated this agent in ``headless'' mode in which the agent executes the task without any user input until the LLM signals task completion to the agent, or the agent detects a loop or reaches a maximum number of steps (250). We applied postprocessing to absolute filepaths to make them compatible with the virtual machine evaluation environment, since the OpenHands agent is run inside its own Docker container.


\section{Detailed Experimental Setup}
\label{app:experiment-setup}

\begin{table}[h]
    \centering
    \footnotesize
    \renewcommand{\arraystretch}{1.1}
    \setlength{\tabcolsep}{11pt}
    \resizebox{0.85\linewidth}{!}{%
    \begin{tabular}{lcc} 
    \toprule
    \textbf{Task} & \textbf{Instance Type} & \makecell{\textbf{Runtime Duration} \\ \textbf{(Gold Solution)}} \\
    \midrule
    CheckEval & CPU & 1m \\
    COGS & K80 & 5h \\
    Entity Tracking & A100 & 2h \\    
    Explain then Translate & CPU & <1m \\
    Instruction Tuning & A100 & 5h \\
    Mission Impossible & A100 & 4h \\
    Othello & K80 & 1h \\
    Reasoning or Reciting & A100 & 6h \\
    Re-reading & A100 & 30m \\
    Tree of Thoughts & A100 & 20m \\
    VariErr-NLI & A100 & 10m \\
    WinoDict & A100 & 30m \\
    \bottomrule
    \end{tabular}}
    \caption{Resource requirements for each task.}
    \label{tab:task-vm-info}
\end{table}
\Cref{tab:task-vm-info} shows the details about the execution environment for each task. 

\section{Solution Complexity}
\begin{table}[ht]
    \centering
    \footnotesize
    \renewcommand{\arraystretch}{1.1}
    \setlength{\tabcolsep}{11pt}
    \resizebox{0.85\linewidth}{!}{%
    \begin{tabular}{lc} 
    \toprule
    \textbf{Task} & \textbf{Delta Cyclomatic Complexity} \\
    \midrule
    CheckEval & 93.0 \\
    COGS & 3.0 \\
    Entity Tracking & 1.0 \\    
    Explain then Translate & 12.0 \\
    Instruction Tuning & 29.0 \\
    Mission Impossible & 19.0 \\
    Othello & 8.0 \\
    Reasoning or Reciting & 16.0 \\
    Re-reading & 2.0 \\
    Tree of Thoughts & 18.0 \\
    VariErr-NLI & 25.0 \\
    WinoDict & 28.0 \\
    \bottomrule
    \end{tabular}}
    \caption{Total net change (delta) in cyclomatic complexity between the original repository state and the gold implementation, computed over the targeted (modified) files.}
    \label{tab:cyc-complexity}
\end{table}
\label{appendix:cyclomatic_complexity}
\Cref{tab:cyc-complexity} shows the total net change (delta) in cyclomatic complexity between the original repository state and the gold implementation, computed only over the files modified in the gold implementation. Cyclomatic complexity was measured using \href{https://github.com/rubik/radon}{radon} for Python and \href{https://github.com/shellspec/shellmetrics}{shellmetrics} for bash. Delta cyclomatic complexity is strongly correlated with lines of code changed (Pearson $r = 0.845$).

\begin{figure*}[ht]
    \centering
    \includegraphics[width=\linewidth]{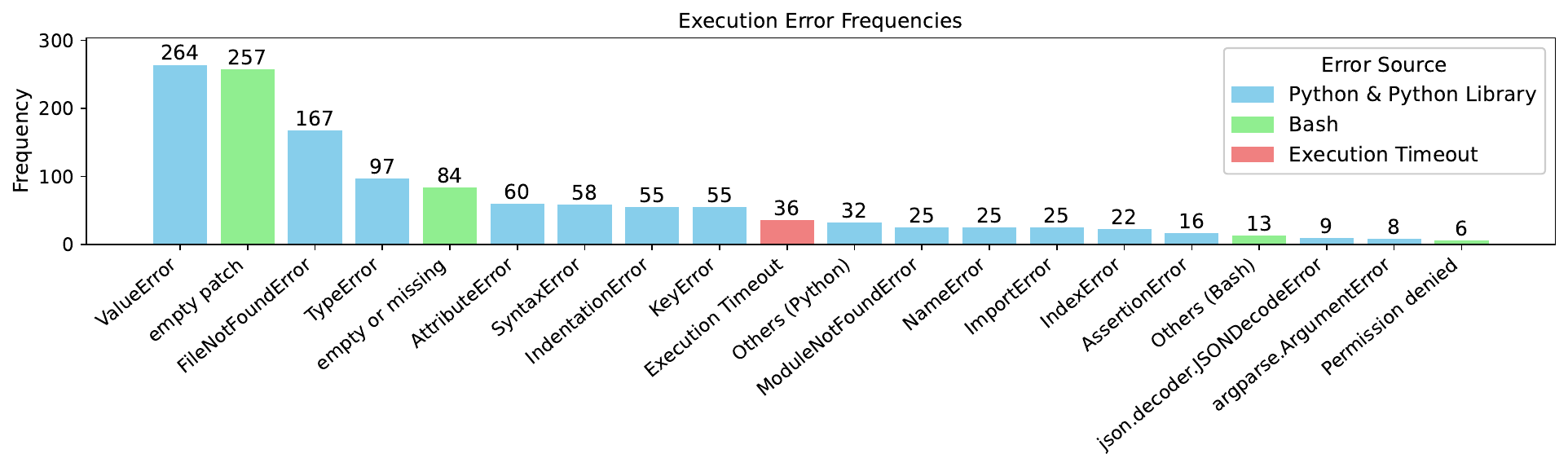}
    \caption{Distribution of execution errors across Python, Bash, and timeout categories. Errors with fewer than 5 occurrences are grouped as `Others'.}
    \label{fig:explicit_error_types}
\end{figure*}

\begin{figure*}[ht]
    \centering
    \includegraphics[width=\linewidth]{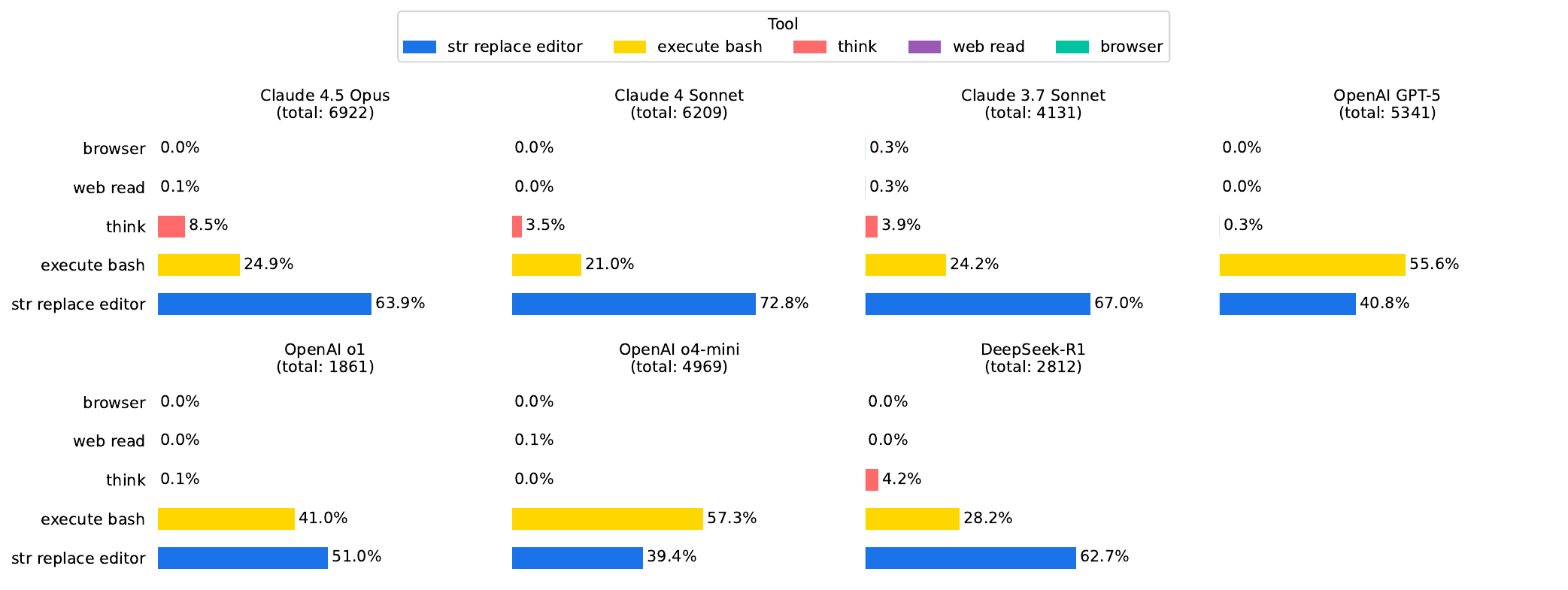}
    \caption{Tool usage distribution across OpenHands agent implementations. Percentages indicate the frequency of each tool type, while the total usage count is shown in each column header.}
    \label{fig:tool_usage}
\end{figure*}

\begin{figure*}[ht]
    \centering
    \includegraphics[width=\linewidth]{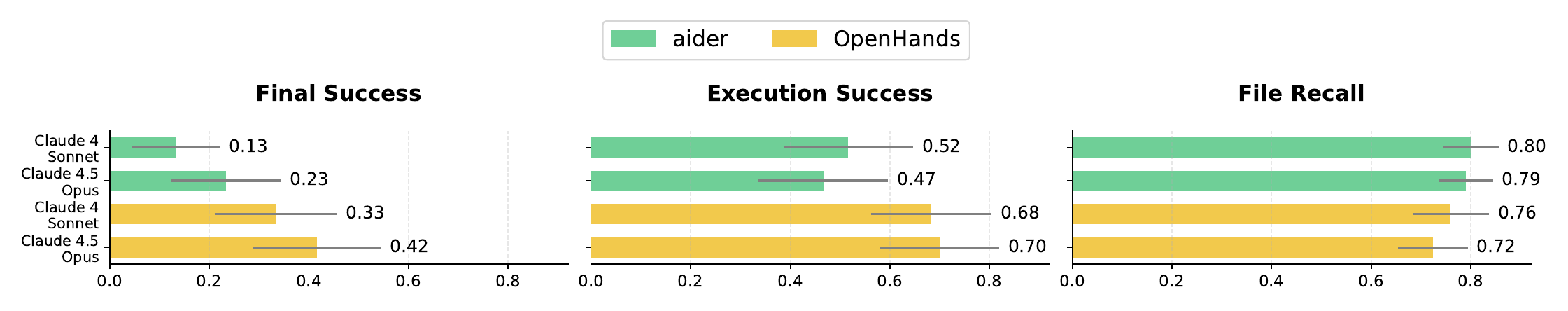}
    \caption{Additional performance results using Claude 4.5 Opus as the backbone LLM.}
    \label{fig:main_results_appendix}
\end{figure*}

\begin{figure}[ht]
    \centering
    \includegraphics[width=0.95\columnwidth]{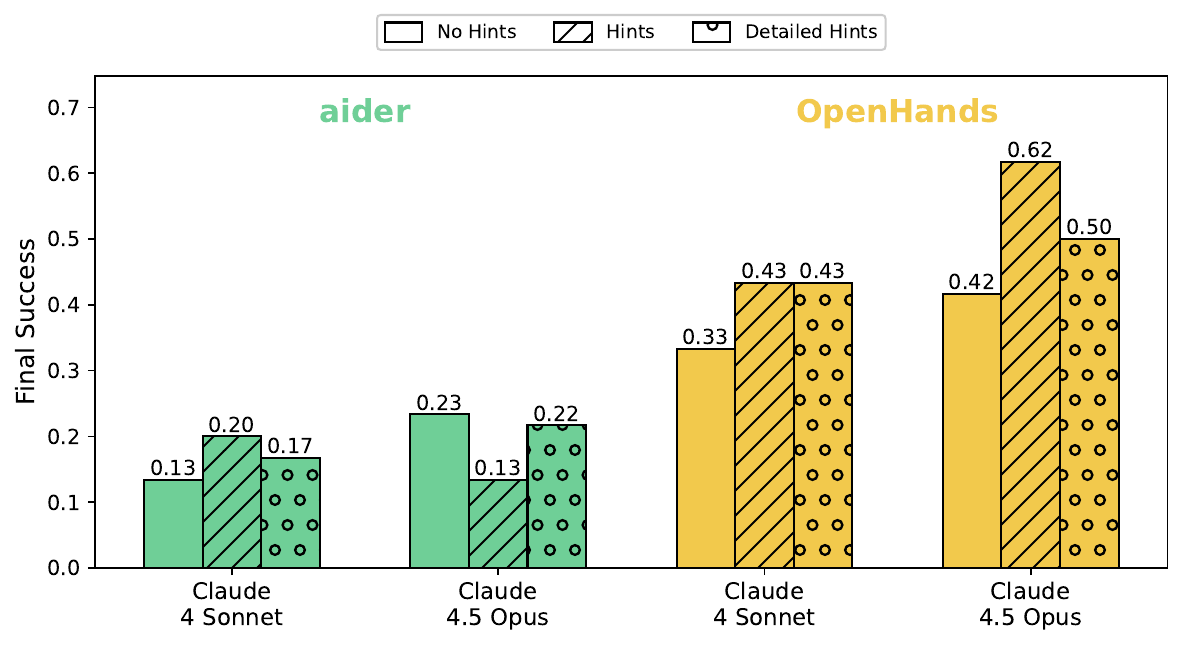}
    \caption{Final success rates for each hint level using Claude 4.5 Opus as the backbone LLM.}
    \label{fig:hint_results_appendix}
\end{figure}

\begin{figure}[ht]
    \centering
    \includegraphics[width=\columnwidth]{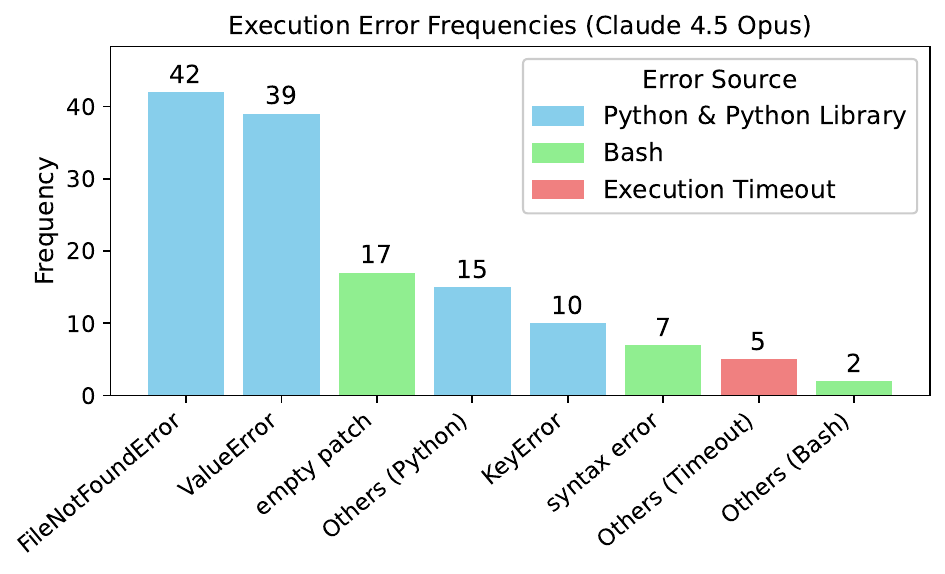}
    \caption{Distribution of execution errors for Claude 4.5 Opus only, categorized into Python, Bash, and timeout errors. Both Aider and OpenHands results are combined. Errors with fewer than 5 occurrences are grouped as `Others'.}
    \label{fig:explicit_error_types_appendix}
\end{figure}

\section{An Example Task Instruction (Extension of WinoDict)}
\label{appendix:winodict_instructions}

\Cref{fig:winodict_instructions} shows the full task instruction for the WinoDict extension.

\begin{figure*}[ht]
\centering
\begin{tcolorbox}[
  enhanced,
  sharp corners,
  colback=white,
  colframe=black,
  title=WinoDict Task Instruction,
  fonttitle=\bfseries,
  boxrule=0.5pt,
  width=\textwidth
]
\footnotesize
\input{winodict_instructions.tex}
\end{tcolorbox}
\caption{WinoDict task instruction.}
\label{fig:winodict_instructions}
\end{figure*}

\section{Tool use/action distribution}
\label{appendix:openhands_tool_usage}
OpenHands agents interact with external tools during execution, and we analyze how their tool usage varies across different LLMs. Claude 4 Sonnet and OpenAI GPT-5 showed the highest overall tool usage. File operations (\texttt{str\_replace\_editor}) and bash commands (\texttt{execute\_bash}) were the most frequently used tools across all models (see \Cref{fig:tool_usage}) but occasionally the agent did also perform web searches or use a browser to render the paper PDF.

\section{Detailed Experimental Results}
\label{app:experiment-details}

\Cref{tab:results-with-hintlevel} shows the detailed results for all metrics and each agent-LLM combination across all three hint levels.

\Cref{tab:token-usage} shows the number of turns as well as the number of input and output tokens, averaged across the three runs for each agent.

\Cref{tab:cost-duration} shows the costs and duration for running each agent on a single task on average, as well as the total cost and total durations, based on the main experiment only (providing no hints). Including preliminary and failed runs not reported in the main paper, we estimate that the total compute required for the full project was approximately 4–5x the reported amount.

\Cref{tab:error_breakdown_1} and \Cref{tab:error_breakdown_2} show the detailed breakdown of errors for each agent and LLM combination.

\Cref{tab:detailed-aider_claude,tab:detailed-aider_claude4,tab:detailed-aider_gpt5,tab:detailed-aider_o1,tab:detailed-aider_o4-mini,tab:detailed-aider_deepseek,tab:detailed-openhands_claude,tab:detailed-openhands_claude_4,tab:detailed-openhands_gpt5,tab:detailed-openhands_o1,tab:detailed-openhands_o4-mini,tab:detailed-openhands_deepseek-r1,tab:detailed-aider_claude_4.5_opus,tab:detailed-openhands_claude_4.5_opus} show the detailed breakdown of task specific performance for each agent and LLM combination.

\begin{table*}[h]
\centering
\resizebox{\linewidth}{!}{%
\renewcommand{\arraystretch}{1.1}
\setlength{\tabcolsep}{8pt}
\footnotesize
\begin{tabular}{lllccc}
\toprule
\textbf{Agent} & \textbf{Model} & \textbf{Hint Level} &
{\textbf{Final Success}} &
{\textbf{Execution Success}} &
{\textbf{File Recall}} \\
\midrule
\multirow[t]{21}{*}{Aider}
& \multirow[t]{3}{*}{Claude 4.5 Opus}
& No hints        & 0.23 & 0.47 & 0.79 \\
&                & Hints           & 0.13 & 0.30 & 0.56 \\
&                & Detailed Hints  & 0.22 & 0.37 & 0.86 \\
& \multirow[t]{3}{*}{Claude 4 Sonnet}
& No hints        & 0.13 & 0.52 & 0.80 \\
&                & Hints           & 0.20 & 0.47 & 0.78 \\
&                & Detailed Hints  & 0.17 & 0.42 & 0.83 \\
& \multirow[t]{3}{*}{Claude 3.7 Sonnet}
& No hints        & 0.08 & 0.23 & 0.81 \\
&                & Hints           & 0.10 & 0.18 & 0.79 \\
&                & Detailed Hints  & 0.05 & 0.20 & 0.81 \\
& \multirow[t]{3}{*}{OpenAI GPT-5}
& No hints        & 0.02 & 0.42 & 0.71 \\
&                & Hints           & 0.00 & 0.38 & 0.74 \\
&                & Detailed Hints  & 0.08 & 0.33 & 0.74 \\
& \multirow[t]{3}{*}{OpenAI o1}
& No hints        & 0.00 & 0.20 & 0.77 \\
&                & Hints           & 0.00 & 0.28 & 0.78 \\
&                & Detailed Hints  & 0.02 & 0.35 & 0.77 \\
& \multirow[t]{3}{*}{OpenAI o4-mini}
& No hints        & 0.03 & 0.20 & 0.45 \\
&                & Hints           & 0.20 & 0.27 & 0.47 \\
&                & Detailed Hints  & 0.17 & 0.33 & 0.51 \\
& \multirow[t]{3}{*}{DeepSeek-R1}
& No hints        & 0.00 & 0.00 & 0.11 \\
&                & Hints           & 0.00 & 0.00 & 0.05 \\
&                & Detailed Hints  & 0.00 & 0.00 & 0.07 \\
\midrule
\multirow[t]{21}{*}{OpenHands}
& \multirow[t]{3}{*}{Claude 4.5 Opus}
& No hints        & 0.42 & 0.70 & 0.72 \\
&                & Hints           & 0.62 & 0.80 & 0.80 \\
&                & Detailed Hints  & 0.50 & 0.82 & 0.84 \\
& \multirow[t]{3}{*}{Claude 4 Sonnet}
& No hints        & 0.33 & 0.68 & 0.76 \\
&                & Hints           & 0.43 & 0.68 & 0.83 \\
&                & Detailed Hints  & 0.43 & 0.78 & 0.89 \\
& \multirow[t]{3}{*}{Claude 3.7 Sonnet}
& No hints        & 0.27 & 0.45 & 0.71 \\
&                & Hints           & 0.40 & 0.53 & 0.81 \\
&                & Detailed Hints  & 0.35 & 0.52 & 0.87 \\
& \multirow[t]{3}{*}{OpenAI GPT-5}
& No hints        & 0.27 & 0.62 & 0.76 \\
&                & Hints           & 0.37 & 0.63 & 0.85 \\
&                & Detailed Hints  & 0.43 & 0.70 & 0.84 \\
& \multirow[t]{3}{*}{OpenAI o1}
& No hints        & 0.00 & 0.33 & 0.56 \\
&                & Hints           & 0.08 & 0.33 & 0.64 \\
&                & Detailed Hints  & 0.03 & 0.38 & 0.73 \\
& \multirow[t]{3}{*}{OpenAI o4-mini}
& No hints        & 0.08 & 0.37 & 0.64 \\
&                & Hints           & 0.22 & 0.37 & 0.73 \\
&                & Detailed Hints  & 0.13 & 0.48 & 0.71 \\
& \multirow[t]{3}{*}{DeepSeek-R1}
& No hints        & 0.00 & 0.17 & 0.52 \\
&                & Hints           & 0.00 & 0.13 & 0.60 \\
&                & Detailed Hints  & 0.10 & 0.22 & 0.68 \\
\bottomrule
\end{tabular}}
\caption{Detailed performance on \benchmark{}, evaluated across three hint levels. Results are averaged across five runs.}
\label{tab:results-with-hintlevel}
\end{table*}

\begin{table*}[t]
\centering
\footnotesize
\resizebox{\textwidth}{!}{%
\renewcommand{\arraystretch}{1.1}
\setlength{\tabcolsep}{11pt}
\begin{tabular}{lll
S[table-format=2.2]
S[table-format=7.2, group-separator={,}]
S[table-format=5.02, group-separator={,}]}
\toprule
\textbf{Agent} & \textbf{Model} & \textbf{Hints Level} &
{\makecell{\textbf{Total Turns}\\\textbf{(Avg.)}}} &
{\makecell{\textbf{Prompt Tokens}\\\textbf{(Avg.)}}} &
{\makecell{\textbf{Output Tokens}\\\textbf{(Avg.)}}} \\
\midrule
\multirow[t]{21}{*}{aider}
& \multirow[t]{3}{*}{Claude 4.5 Opus}
& No hints       & 2.00 & 3398.40 & 4176.70 \\
&                & Hints          & 2.00 & 3823.90 & 2663.80 \\
&                & Detailed Hints & 2.00 & 4285.80 & 3804.80 \\
& \multirow[t]{3}{*}{Claude 4 Sonnet}
& No hints       & 2.00 & 2941.70 & 4048.60 \\
&                & Hints          & 2.00 & 3028.20 & 4275.00 \\
&                & Detailed Hints & 2.00 & 3412.40 & 4187.10 \\
& \multirow[t]{3}{*}{Claude 3.7 Sonnet}
& No hints       & 2.00 & 2941.80 & 4669.20 \\
&                & Hints          & 2.00 & 3029.10 & 4021.70 \\
&                & Detailed Hints & 2.00 & 3413.00 & 3950.10 \\
& \multirow[t]{3}{*}{OpenAI GPT-5}
& No hints       & 2.00 & 2881.20 & 8650.00 \\
&                & Hints          & 2.00 & 2976.60 & 8721.40 \\
&                & Detailed Hints & 2.00 & 3360.50 & 9650.00 \\
& \multirow[t]{3}{*}{OpenAI o1}
& No hints       & 2.00 & 2964.70 & 5536.30 \\
&                & Hints          & 2.00 & 3061.80 & 6086.70 \\
&                & Detailed Hints & 2.00 & 3447.50 & 6088.30 \\
& \multirow[t]{3}{*}{OpenAI o4-mini}
& No hints       & 2.00 & 2910.40 & 3944.60 \\
&                & Hints          & 2.00 & 3002.40 & 3860.30 \\
&                & Detailed Hints & 2.00 & 3389.30 & 4648.10 \\
& \multirow[t]{3}{*}{DeepSeek-R1}
& No hints       & 2.00 & 2966.10 & 3732.20 \\
&                & Hints          & 2.00 & 3002.60 & 3850.90 \\
&                & Detailed Hints & 2.00 & 3446.20 & 3570.60 \\
\midrule
\multirow[t]{21}{*}{OpenHands}
& \multirow[t]{3}{*}{Claude 4.5 Opus}
& No hints       & 87.68 & 1645781.35 & 12550.75 \\
&                & Hints          & 78.92 & 1392933.87 & 12215.67 \\
&                & Detailed Hints & 68.97 & 1140871.72 & 11380.53 \\
& \multirow[t]{3}{*}{Claude 4 Sonnet}
& No hints       & 99.33 & 1847555.70 & 9872.00 \\
&                & Hints          & 97.12 & 1822132.70 & 9791.90 \\
&                & Detailed Hints & 95.33 & 1390346.00 & 9814.20 \\
& \multirow[t]{3}{*}{Claude 3.7 Sonnet}
& No hints       & 44.90 & 476605.30  & 6932.30 \\
&                & Hints          & 45.18 & 488315.90  & 6850.60 \\
&                & Detailed Hints & 51.68 & 561399.10  & 7344.30 \\
& \multirow[t]{3}{*}{OpenAI GPT-5}
& No hints       & 69.62 & 871624.30  & 37786.20 \\
&                & Hints          & 71.63 & 1010020.60 & 40139.40 \\
&                & Detailed Hints & 67.30 & 947675.20  & 34107.50 \\
& \multirow[t]{3}{*}{OpenAI o1}
& No hints       & 26.12 & 164538.00  & 18007.30 \\
&                & Hints          & 22.73 & 131386.80  & 12660.40 \\
&                & Detailed Hints & 17.30 & 92010.20   & 10151.30 \\
& \multirow[t]{3}{*}{OpenAI o4-mini}
& No hints       & 59.53 & 617804.50  & 28767.90 \\
&                & Hints          & 58.95 & 592439.80  & 27485.30 \\
&                & Detailed Hints & 52.60 & 520614.00  & 24594.10 \\
& \multirow[t]{3}{*}{DeepSeek-R1}
& No hints       & 36.10 & 270543.70  & 18085.10 \\
&                & Hints          & 35.93 & 220510.30  & 17681.90 \\
&                & Detailed Hints & 35.72 & 211106.10  & 16938.80 \\
\bottomrule
\end{tabular}}
\caption{Token usage statistics across agents and models.}
\label{tab:token-usage}
\end{table*}

\begin{table*}[ht]
    \centering
    \footnotesize
    \renewcommand{\arraystretch}{1.1}
    \setlength{\tabcolsep}{11pt}
    \sisetup{
        table-align-text-post = false,
        detect-weight = true,
        detect-family = true,
        round-mode = places,
        round-precision = 2,
    }
    \resizebox{\textwidth}{!}{%
        \begin{tabular}{ll
            S[table-format=1.2]
            >{\raggedleft\arraybackslash}p{2cm}
            S[table-format=4.2]
            >{\raggedleft\arraybackslash}p{2cm}} 
            \toprule
            \textbf{Agent} & \textbf{Model} & {\textbf{Avg. Cost (\$)}} & \textbf{Avg. Duration} & {\textbf{Total Cost (\$)}} & \textbf{Total Duration} \\
            \midrule
            \multirow[t]{7}{*}{aider} 
            & Claude 4.5 Opus     & 0.37 & 57s     & 67.35  & 2h 51m 36s \\
            & Claude 4 Sonnet     & 0.20 & 1m 10s  & 35.13  & 3h 28m 38s \\
            & Claude 3.7 Sonnet   & 0.26 & 1m 18s  & 45.99  & 3h 54m 57s \\
            & OpenAI GPT-5        & 0.11 & 3m 6s   & 19.47  & 9h 16m 45s \\
            & OpenAI o1           & 0.69 & 2m 21s  & 124.69 & 7h 1m 34s \\
            & OpenAI o4-mini      & 0.03 & 49s     & 5.45   & 2h 27m 25s \\
            & DeepSeek-R1         & 0.02 & 1m 29s  & 3.94   & 4h 22m 7s \\
            \midrule
            \multirow[t]{7}{*}{OpenHands} 
            & Claude 4.5 Opus     & 1.34 & 4m 22s  & 241.26 & 13h 8m 9s \\
            & Claude 4 Sonnet     & 5.21 & 6m 12s  & 937.34 & 18h 38m 42s \\
            & Claude 3.7 Sonnet   & 0.37 & 2m 33s  & 67.20  & 7h 41m 29s \\
            & OpenAI GPT-5        & 1.16 & 13m 58s & 208.07 & 41h 56m 30s \\
            & OpenAI o1           & 1.91 & 3m 14s  & 343.44 & 9h 42m 56s \\
            & OpenAI o4-mini      & 0.55 & 5m 30s  & 98.97  & 16h 31m 49s \\
            & DeepSeek-R1         & 0.04 & 13m 22s & 7.99   & 40h 8m 16s \\
            \bottomrule
        \end{tabular}}    
    \caption{Cost and duration statistics across agents and models (main experiment).}
    \label{tab:cost-duration}
\end{table*}

\begin{table*}[ht]
    \centering
    \footnotesize
    \renewcommand{\arraystretch}{1.1}
    \setlength{\tabcolsep}{8pt}
    \begin{tabular}{lcc} 
    \toprule
    \textbf{Task} & \textbf{Repository} & \textbf{License} \\
    \midrule
    CheckEval & yukyunglee/CheckEval & MIT \\
    COGS & najoungkim/COGS & MIT \\
    Entity Tracking & najoungkim/code-models-entity-tracking & Apache 2.0 \\    
    Explain then Translate & PootieT/explain-then-translate & MIT \\
    Instruction Tuning & john-hewitt/implicit-ins & Apache 2.0 \\
    Mission Impossible & jkallini/mission-impossible-language-models & ??? \\
    Othello & likenneth/othello\_world & MIT \\
    Reasoning or Reciting & ZhaofengWu/counterfactual-evaluation & MIT \\
    Re-reading & EleutherAI/lm-evaluation-harness & MIT \\
    Tree of Thoughts & princeton-nlp/tree-of-thought-llm & MIT \\
    VariErr-NLI & mainlp/VariErr-NLI & MIT \\
    WinoDict & google-research/language/tree/master/language/wino\_dict & Apache 2.0 \\
    \bottomrule
    \end{tabular}
    \caption{Licenses for each Github repository.}
    \label{tab:task-license-info}
\end{table*}

\begin{table*}[h]
\centering
\setlength{\tabcolsep}{11pt}
\resizebox{\linewidth}{!}{%
\begin{tabular}{lccccccc}
\toprule
\textbf{Error Type} & \multicolumn{7}{c}{\textbf{Aider}} \\
\cmidrule(lr){2-8}
& \begin{tabular}[c]{@{}c@{}}Claude\\4.5 Opus\end{tabular} & \begin{tabular}[c]{@{}c@{}}Claude\\4 Sonnet\end{tabular} & \begin{tabular}[c]{@{}c@{}}Claude\\3.7 Sonnet\end{tabular} & \begin{tabular}[c]{@{}c@{}}OpenAI\\GPT-5\end{tabular} & \begin{tabular}[c]{@{}c@{}}OpenAI\\o1\end{tabular} & \begin{tabular}[c]{@{}c@{}}OpenAI\\o4-mini\end{tabular} & \begin{tabular}[c]{@{}c@{}}DeepSeek\\R1\end{tabular} \\
\midrule
\multicolumn{8}{l}{\textit{\textbf{Python Errors}}} \\
\midrule
AssertionError & 0 & 2 & 0 & 5 & 0 & 2 & 0 \\
AttributeError & 1 & 6 & 11 & 6 & 9 & 3 & 0 \\
FileNotFoundError & 41 & 21 & 73 & 11 & 19 & 7 & 3 \\
ImportError & 0 & 1 & 6 & 3 & 10 & 0 & 0 \\
IndentationError & 0 & 0 & 0 & 0 & 0 & 0 & 0 \\
IndexError & 2 & 0 & 0 & 3 & 2 & 0 & 0 \\
IsADirectoryError & 0 & 0 & 0 & 0 & 0 & 0 & 0 \\
KeyError & 3 & 1 & 5 & 2 & 3 & 12 & 2 \\
LookupError & 0 & 0 & 0 & 1 & 0 & 0 & 0 \\
ModuleNotFoundError & 0 & 2 & 0 & 7 & 4 & 0 & 0 \\
NameError & 0 & 1 & 1 & 1 & 1 & 0 & 1 \\
NotImplementedError & 0 & 0 & 0 & 0 & 0 & 0 & 0 \\
OSError & 0 & 0 & 0 & 0 & 0 & 0 & 0 \\
RuntimeError & 0 & 0 & 0 & 2 & 2 & 0 & 0 \\
SyntaxError & 1 & 1 & 0 & 0 & 10 & 6 & 0 \\
TypeError & 2 & 5 & 4 & 7 & 3 & 11 & 6 \\
UnboundLocalError & 0 & 2 & 0 & 0 & 2 & 0 & 0 \\
ValueError & 23 & 39 & 37 & 35 & 18 & 10 & 7 \\
EOFError & 0 & 0 & 1 & 0 & 0 & 0 & 0 \\
\midrule
\multicolumn{8}{l}{\textit{\textbf{Python Library Errors}}} \\
\midrule
DatasetNotFoundError & 0 & 1 & 0 & 0 & 0 & 0 & 0 \\
NotFoundError & 1 & 2 & 0 & 1 & 2 & 0 & 0 \\
OutOfMemory & 0 & 0 & 0 & 0 & 0 & 0 & 0 \\
ArgumentError & 0 & 0 & 0 & 0 & 0 & 0 & 0 \\
ScannerError & 0 & 0 & 0 & 0 & 0 & 0 & 0 \\
\midrule
\multicolumn{8}{l}{\textit{\textbf{Other Python Errors}}} \\
\midrule
ConstructorError & 0 & 2 & 0 & 0 & 0 & 0 & 0 \\
JSONDecodeError & 0 & 3 & 0 & 5 & 0 & 0 & 0 \\
HFValidationError & 1 & 0 & 0 & 0 & 0 & 0 & 0 \\
ParserError & 0 & 0 & 0 & 0 & 0 & 0 & 0 \\
\midrule
\multicolumn{8}{l}{\textit{\textbf{Bash Errors}}} \\
\midrule
cannot create directory & 0 & 0 & 0 & 0 & 1 & 0 & 0 \\
empty patch & 17 & 5 & 0 & 13 & 0 & 69 & 111 \\
empty or missing & 0 & 0 & 0 & 5 & 7 & 7 & 41 \\
unable to write file & 0 & 0 & 0 & 0 & 0 & 0 & 5 \\
Permission denied & 0 & 0 & 0 & 0 & 3 & 0 & 0 \\
syntax error & 0 & 0 & 0 & 0 & 0 & 0 & 0 \\
cannot access & 0 & 0 & 0 & 0 & 0 & 0 & 0 \\
patch failed & 2 & 1 & 0 & 0 & 0 & 0 & 0 \\
\midrule
\textit{\textbf{Execution Timeout}} & 1 & 0 & 2 & 0 & 12 & 1 & 2 \\
\bottomrule
\end{tabular}}
\caption{Breakdown of error counts for Aider.}
\label{tab:error_breakdown_1}
\end{table*}

\begin{table*}[h]
\centering
\setlength{\tabcolsep}{11pt}
\resizebox{\linewidth}{!}{%
\begin{tabular}{lccccccc}
\toprule
\textbf{Error Type} & \multicolumn{7}{c}{\textbf{OpenHands}} \\
\cmidrule(lr){2-8}
& \begin{tabular}[c]{@{}c@{}}Claude\\4.5 Opus\end{tabular} & \begin{tabular}[c]{@{}c@{}}Claude\\4 Sonnet\end{tabular} & \begin{tabular}[c]{@{}c@{}}Claude\\3.7 Sonnet\end{tabular} & \begin{tabular}[c]{@{}c@{}}OpenAI\\GPT-5\end{tabular} & \begin{tabular}[c]{@{}c@{}}OpenAI\\o1\end{tabular} & \begin{tabular}[c]{@{}c@{}}OpenAI\\o4-mini\end{tabular} & \begin{tabular}[c]{@{}c@{}}DeepSeek\\R1\end{tabular} \\
\midrule
\multicolumn{8}{l}{\textit{\textbf{Python Errors}}} \\
\midrule
AssertionError & 0 & 0 & 0 & 1 & 3 & 3 & 0 \\
AttributeError & 0 & 0 & 10 & 1 & 3 & 3 & 8 \\
FileNotFoundError & 1 & 6 & 11 & 1 & 4 & 2 & 7 \\
ImportError & 0 & 0 & 0 & 0 & 0 & 1 & 4 \\
IndentationError & 0 & 0 & 0 & 4 & 18 & 14 & 19 \\
IndexError & 1 & 6 & 2 & 5 & 1 & 0 & 2 \\
IsADirectoryError & 1 & 0 & 0 & 0 & 0 & 0 & 2 \\
KeyError & 7 & 1 & 6 & 1 & 2 & 2 & 14 \\
LookupError & 0 & 0 & 0 & 1 & 0 & 0 & 0 \\
ModuleNotFoundError & 1 & 4 & 4 & 1 & 0 & 2 & 1 \\
NameError & 0 & 0 & 0 & 2 & 8 & 4 & 6 \\
NotImplementedError & 0 & 0 & 0 & 0 & 0 & 2 & 0 \\
OSError & 0 & 0 & 0 & 0 & 0 & 0 & 2 \\
RuntimeError & 2 & 0 & 0 & 0 & 0 & 0 & 0 \\
SyntaxError & 0 & 0 & 0 & 1 & 21 & 12 & 7 \\
TypeError & 0 & 7 & 24 & 4 & 4 & 4 & 13 \\
UnboundLocalError & 0 & 0 & 0 & 0 & 0 & 0 & 0 \\
ValueError & 16 & 15 & 26 & 33 & 10 & 5 & 11 \\
EOFError & 0 & 0 & 0 & 0 & 0 & 0 & 0 \\
\midrule
\multicolumn{8}{l}{\textit{\textbf{Python Library Errors}}} \\
\midrule
DatasetNotFoundError & 0 & 0 & 0 & 0 & 0 & 0 & 0 \\
NotFoundError & 0 & 0 & 0 & 0 & 1 & 0 & 0 \\
OutOfMemory & 0 & 0 & 0 & 1 & 0 & 3 & 0 \\
ArgumentError & 0 & 0 & 0 & 0 & 0 & 8 & 0 \\
ScannerError & 0 & 0 & 0 & 0 & 0 & 0 & 0 \\
\midrule
\multicolumn{8}{l}{\textit{\textbf{Other Python Errors}}} \\
\midrule
ConstructorError & 0 & 0 & 0 & 0 & 0 & 0 & 0 \\
JSONDecodeError & 0 & 0 & 0 & 1 & 0 & 0 & 0 \\
HFValidationError & 0 & 0 & 0 & 1 & 0 & 0 & 0 \\
ParserError & 0 & 0 & 0 & 1 & 0 & 0 & 0 \\
\midrule
\multicolumn{8}{l}{\textit{\textbf{Bash Errors}}} \\
\midrule
cannot create directory & 0 & 0 & 0 & 0 & 0 & 0 & 0 \\
empty patch & 0 & 4 & 0 & 1 & 15 & 17 & 22 \\
empty or missing & 0 & 0 & 2 & 0 & 5 & 10 & 7 \\
unable to write file & 0 & 0 & 0 & 0 & 0 & 0 & 0 \\
Permission denied & 0 & 0 & 0 & 0 & 2 & 0 & 1 \\
syntax error & 7 & 4 & 0 & 0 & 0 & 0 & 0 \\
cannot access & 0 & 0 & 1 & 0 & 0 & 0 & 0 \\
patch failed & 0 & 0 & 0 & 0 & 0 & 0 & 0 \\
\midrule
\textit{\textbf{Execution Timeout}} & 4 & 0 & 0 & 2 & 8 & 3 & 2 \\
\bottomrule
\end{tabular}}
\caption{Breakdown of error counts for OpenHands.}
\label{tab:error_breakdown_2}
\end{table*}

\begin{table*}
\centering
\resizebox{\linewidth}{!}{%
\renewcommand{\arraystretch}{1.3}
\footnotesize
\begin{tabular}{llllccc}
\toprule
\textbf{Agent} & \textbf{Model} & \textbf{Hint Level} & \textbf{Task} & \textbf{Final Success} & \textbf{Execution Success} & \textbf{File Recall} \\
\midrule
\multirow[t]{36}{*}{aider} & \multirow[t]{36}{*}{Claude 4 Sonnet} & \multirow[t]{12}{*}{No Hints}
  & CheckEval & 0.00 & 1.00 & 1.00 \\
  & & & COGS & 0.00 & 0.00 & 0.50 \\
  & & & Entity Tracking & 0.60 & 0.80 & 1.00 \\
  & & & Explain then Translate & 0.00 & 1.00 & 1.00 \\
  & & & Instruction Tuning & 0.00 & 0.00 & 0.00 \\
  & & & Mission Impossible & 0.00 & 1.00 & 1.00 \\
  & & & Othello & 1.00 & 1.00 & 1.00 \\
  & & & Re-reading & 0.00 & 0.00 & 0.67 \\
  & & & Reasoning or Reciting & 0.00 & 0.80 & 0.52 \\
  & & & Tree Of Thoughts & 0.00 & 0.00 & 0.67 \\
  & & & VariErr-NLI & 0.00 & 0.00 & 1.00 \\
  & & & WinoDict & 0.00 & 0.60 & 0.75 \\
\cdashline{3-7}
  & & \multirow[t]{12}{*}{Hints}  & CheckEval & 0.00 & 1.00 & 1.00 \\
  & & & COGS & 0.20 & 0.20 & 0.50 \\
  & & & Entity Tracking & 0.60 & 0.80 & 1.00 \\
  & & & Explain then Translate & 0.00 & 0.40 & 0.80 \\
  & & & Instruction Tuning & 0.00 & 0.00 & 0.00 \\
  & & & Mission Impossible & 0.80 & 1.00 & 1.00 \\
  & & & Othello & 0.80 & 0.80 & 1.00 \\
  & & & Re-reading & 0.00 & 0.60 & 0.67 \\
  & & & Reasoning or Reciting & 0.00 & 0.40 & 0.60 \\
  & & & Tree Of Thoughts & 0.00 & 0.00 & 0.67 \\
  & & & VariErr-NLI & 0.00 & 0.20 & 0.90 \\
  & & & WinoDict & 0.00 & 0.20 & 0.75 \\
\cdashline{3-7}
  & & \multirow[t]{12}{*}{Detailed Hints}  & CheckEval & 0.00 & 0.80 & 1.00 \\
  & & & COGS & 0.00 & 0.00 & 1.00 \\
  & & & Entity Tracking & 0.00 & 0.20 & 1.00 \\
  & & & Explain then Translate & 0.80 & 1.00 & 1.00 \\
  & & & Instruction Tuning & 0.00 & 0.00 & 0.00 \\
  & & & Mission Impossible & 0.40 & 1.00 & 1.00 \\
  & & & Othello & 0.80 & 1.00 & 1.00 \\
  & & & Re-reading & 0.00 & 0.00 & 0.67 \\
  & & & Reasoning or Reciting & 0.00 & 0.40 & 0.60 \\
  & & & Tree Of Thoughts & 0.00 & 0.00 & 0.53 \\
  & & & VariErr-NLI & 0.00 & 0.00 & 1.00 \\
  & & & WinoDict & 0.00 & 0.60 & 0.75 \\
\bottomrule
\end{tabular}}
\caption{Detailed performance on aider + Claude 4 Sonnet.}
\label{tab:detailed-aider_claude4}
\end{table*}

\begin{table*}
\centering
\resizebox{\linewidth}{!}{%
\renewcommand{\arraystretch}{1.3}
\footnotesize
\begin{tabular}{llllccc}
\toprule
\textbf{Agent} & \textbf{Model} & \textbf{Hint Level} & \textbf{Task} & \textbf{Final Success} & \textbf{Execution Success} & \textbf{File Recall} \\
\midrule
\multirow[t]{36}{*}{aider} & \multirow[t]{36}{*}{Claude 3.7 Sonnet} & \multirow[t]{12}{*}{No Hints}
  & CheckEval & 0.00 & 0.00 & 1.00 \\
  & & & COGS & 0.00 & 0.00 & 0.50 \\
  & & & Entity Tracking & 0.20 & 0.20 & 1.00 \\
  & & & Explain then Translate & 0.00 & 0.60 & 1.00 \\
  & & & Instruction Tuning & 0.00 & 0.40 & 0.00 \\
  & & & Mission Impossible & 0.20 & 0.60 & 1.00 \\
  & & & Othello & 0.60 & 0.60 & 1.00 \\
  & & & Re-reading & 0.00 & 0.20 & 0.67 \\
  & & & Reasoning or Reciting & 0.00 & 0.00 & 0.56 \\
  & & & Tree Of Thoughts & 0.00 & 0.00 & 0.67 \\
  & & & VariErr-NLI & 0.00 & 0.00 & 1.00 \\
  & & & WinoDict & 0.00 & 0.20 & 0.75 \\
\cdashline{3-7}
  & & \multirow[t]{12}{*}{Hints}  & CheckEval & 0.00 & 0.00 & 1.00 \\
  & & & COGS & 0.00 & 0.00 & 0.50 \\
  & & & Entity Tracking & 0.20 & 0.40 & 1.00 \\
  & & & Explain then Translate & 0.00 & 0.60 & 1.00 \\
  & & & Instruction Tuning & 0.00 & 0.00 & 0.00 \\
  & & & Mission Impossible & 0.40 & 0.60 & 1.00 \\
  & & & Othello & 0.60 & 0.60 & 1.00 \\
  & & & Re-reading & 0.00 & 0.00 & 0.67 \\
  & & & Reasoning or Reciting & 0.00 & 0.00 & 0.40 \\
  & & & Tree Of Thoughts & 0.00 & 0.00 & 0.67 \\
  & & & VariErr-NLI & 0.00 & 0.00 & 1.00 \\
  & & & WinoDict & 0.00 & 0.00 & 0.75 \\
\cdashline{3-7}
  & & \multirow[t]{12}{*}{Detailed Hints}  & CheckEval & 0.00 & 0.40 & 1.00 \\
  & & & COGS & 0.00 & 0.00 & 0.60 \\
  & & & Entity Tracking & 0.00 & 0.00 & 1.00 \\
  & & & Explain then Translate & 0.60 & 0.60 & 1.00 \\
  & & & Instruction Tuning & 0.00 & 0.00 & 0.00 \\
  & & & Mission Impossible & 0.00 & 0.60 & 1.00 \\
  & & & Othello & 0.00 & 0.60 & 1.00 \\
  & & & Re-reading & 0.00 & 0.00 & 0.67 \\
  & & & Reasoning or Reciting & 0.00 & 0.00 & 0.44 \\
  & & & Tree Of Thoughts & 0.00 & 0.00 & 0.80 \\
  & & & VariErr-NLI & 0.00 & 0.00 & 1.00 \\
  & & & WinoDict & 0.00 & 0.20 & 0.75 \\
\bottomrule
\end{tabular}}
\caption{Detailed performance on aider + Claude 3.7 Sonnet.}
\label{tab:detailed-aider_claude}
\end{table*}

\begin{table*}
\centering
\resizebox{\linewidth}{!}{%
\renewcommand{\arraystretch}{1.3}
\footnotesize
\begin{tabular}{llllccc}
\toprule
\textbf{Agent} & \textbf{Model} & \textbf{Hint Level} & \textbf{Task} & \textbf{Final Success} & \textbf{Execution Success} & \textbf{File Recall} \\
\midrule
\multirow[t]{36}{*}{aider} & \multirow[t]{36}{*}{OpenAI GPT-5} & \multirow[t]{12}{*}{No Hints}
  & CheckEval & 0.00 & 0.60 & 0.80 \\
  & & & COGS & 0.00 & 0.60 & 1.00 \\
  & & & Entity Tracking & 0.00 & 0.60 & 1.00 \\
  & & & Explain then Translate & 0.00 & 1.00 & 1.00 \\
  & & & Instruction Tuning & 0.00 & 0.00 & 0.00 \\
  & & & Mission Impossible & 0.00 & 0.40 & 1.00 \\
  & & & Othello & 0.00 & 0.80 & 1.00 \\
  & & & Re-reading & 0.00 & 0.20 & 0.13 \\
  & & & Reasoning or Reciting & 0.00 & 0.00 & 0.40 \\
  & & & Tree Of Thoughts & 0.00 & 0.00 & 0.60 \\
  & & & VariErr-NLI & 0.00 & 0.00 & 1.00 \\
  & & & WinoDict & 0.20 & 0.80 & 0.60 \\
\cdashline{3-7}
  & & \multirow[t]{12}{*}{Hints}  & CheckEval & 0.00 & 0.60 & 1.00 \\
  & & & COGS & 0.00 & 0.60 & 1.00 \\
  & & & Entity Tracking & 0.00 & 0.60 & 1.00 \\
  & & & Explain then Translate & 0.00 & 1.00 & 1.00 \\
  & & & Instruction Tuning & 0.00 & 0.00 & 0.00 \\
  & & & Mission Impossible & 0.00 & 0.20 & 1.00 \\
  & & & Othello & 0.00 & 1.00 & 1.00 \\
  & & & Re-reading & 0.00 & 0.20 & 0.13 \\
  & & & Reasoning or Reciting & 0.00 & 0.20 & 0.56 \\
  & & & Tree Of Thoughts & 0.00 & 0.00 & 0.47 \\
  & & & VariErr-NLI & 0.00 & 0.00 & 1.00 \\
  & & & WinoDict & 0.00 & 0.20 & 0.75 \\
\cdashline{3-7}
  & & \multirow[t]{12}{*}{Detailed Hints}  & CheckEval & 0.00 & 0.20 & 1.00 \\
  & & & COGS & 0.00 & 1.00 & 1.00 \\
  & & & Entity Tracking & 0.00 & 0.60 & 0.80 \\
  & & & Explain then Translate & 0.80 & 1.00 & 1.00 \\
  & & & Instruction Tuning & 0.00 & 0.00 & 0.00 \\
  & & & Mission Impossible & 0.20 & 0.60 & 1.00 \\
  & & & Othello & 0.00 & 0.00 & 1.00 \\
  & & & Re-reading & 0.00 & 0.00 & 0.13 \\
  & & & Reasoning or Reciting & 0.00 & 0.40 & 0.48 \\
  & & & Tree Of Thoughts & 0.00 & 0.00 & 0.67 \\
  & & & VariErr-NLI & 0.00 & 0.00 & 1.00 \\
  & & & WinoDict & 0.00 & 0.20 & 0.75 \\
\bottomrule
\end{tabular}}
\caption{Detailed performance on aider + OpenAI GPT-5.}
\label{tab:detailed-aider_gpt5}
\end{table*}

\begin{table*}
\centering
\resizebox{\linewidth}{!}{%
\renewcommand{\arraystretch}{1.3}
\footnotesize
\begin{tabular}{llllccc}
\toprule
\textbf{Agent} & \textbf{Model} & \textbf{Hint Level} & \textbf{Task} & \textbf{Final Success} & \textbf{Execution Success} & \textbf{File Recall} \\
\midrule
\multirow[t]{36}{*}{aider} & \multirow[t]{36}{*}{OpenAI o1} & \multirow[t]{12}{*}{No Hints}
  & CheckEval & 0.00 & 0.40 & 0.80 \\
  & & & COGS & 0.00 & 0.00 & 1.00 \\
  & & & Entity Tracking & 0.00 & 1.00 & 1.00 \\
  & & & Explain then Translate & 0.00 & 0.20 & 1.00 \\
  & & & Instruction Tuning & 0.00 & 0.00 & 0.00 \\
  & & & Mission Impossible & 0.00 & 0.20 & 1.00 \\
  & & & Othello & 0.00 & 0.00 & 1.00 \\
  & & & Re-reading & 0.00 & 0.00 & 0.67 \\
  & & & Reasoning or Reciting & 0.00 & 0.00 & 0.36 \\
  & & & Tree Of Thoughts & 0.00 & 0.00 & 0.67 \\
  & & & VariErr-NLI & 0.00 & 0.00 & 1.00 \\
  & & & WinoDict & 0.00 & 0.60 & 0.75 \\
\cdashline{3-7}
  & & \multirow[t]{12}{*}{Hints}  & CheckEval & 0.00 & 0.80 & 0.80 \\
  & & & COGS & 0.00 & 0.00 & 1.00 \\
  & & & Entity Tracking & 0.00 & 1.00 & 1.00 \\
  & & & Explain then Translate & 0.00 & 0.00 & 1.00 \\
  & & & Instruction Tuning & 0.00 & 0.00 & 0.00 \\
  & & & Mission Impossible & 0.00 & 0.60 & 1.00 \\
  & & & Othello & 0.00 & 0.20 & 1.00 \\
  & & & Re-reading & 0.00 & 0.00 & 0.67 \\
  & & & Reasoning or Reciting & 0.00 & 0.00 & 0.60 \\
  & & & Tree Of Thoughts & 0.00 & 0.60 & 0.67 \\
  & & & VariErr-NLI & 0.00 & 0.00 & 1.00 \\
  & & & WinoDict & 0.00 & 0.20 & 0.65 \\
\cdashline{3-7}
  & & \multirow[t]{12}{*}{Detailed Hints}  & CheckEval & 0.00 & 0.00 & 0.80 \\
  & & & COGS & 0.00 & 0.60 & 1.00 \\
  & & & Entity Tracking & 0.00 & 1.00 & 1.00 \\
  & & & Explain then Translate & 0.20 & 0.20 & 1.00 \\
  & & & Instruction Tuning & 0.00 & 0.00 & 0.00 \\
  & & & Mission Impossible & 0.00 & 0.60 & 1.00 \\
  & & & Othello & 0.00 & 1.00 & 1.00 \\
  & & & Re-reading & 0.00 & 0.20 & 0.67 \\
  & & & Reasoning or Reciting & 0.00 & 0.00 & 0.48 \\
  & & & Tree Of Thoughts & 0.00 & 0.00 & 0.60 \\
  & & & VariErr-NLI & 0.00 & 0.00 & 1.00 \\
  & & & WinoDict & 0.00 & 0.60 & 0.75 \\
\bottomrule
\end{tabular}}
\caption{Detailed performance on aider + OpenAI o1.}
\label{tab:detailed-aider_o1}
\end{table*}

\begin{table*}
\centering
\resizebox{\linewidth}{!}{%
\renewcommand{\arraystretch}{1.3}
\footnotesize
\begin{tabular}{llllccc}
\toprule
\textbf{Agent} & \textbf{Model} & \textbf{Hint Level} & \textbf{Task} & \textbf{Final Success} & \textbf{Execution Success} & \textbf{File Recall} \\
\midrule
\multirow[t]{36}{*}{aider} & \multirow[t]{36}{*}{OpenAI o4-mini} & \multirow[t]{12}{*}{No Hints}
  & CheckEval & 0.00 & 0.00 & 0.60 \\
  & & & COGS & 0.00 & 0.00 & 0.00 \\
  & & & Entity Tracking & 0.00 & 0.60 & 0.80 \\
  & & & Explain then Translate & 0.00 & 0.40 & 0.80 \\
  & & & Instruction Tuning & 0.00 & 0.00 & 0.00 \\
  & & & Mission Impossible & 0.40 & 0.40 & 0.90 \\
  & & & Othello & 0.00 & 0.40 & 1.00 \\
  & & & Re-reading & 0.00 & 0.00 & 0.00 \\
  & & & Reasoning or Reciting & 0.00 & 0.00 & 0.16 \\
  & & & Tree Of Thoughts & 0.00 & 0.40 & 0.60 \\
  & & & VariErr-NLI & 0.00 & 0.00 & 0.20 \\
  & & & WinoDict & 0.00 & 0.20 & 0.30 \\
\cdashline{3-7}
  & & \multirow[t]{12}{*}{Hints}  & CheckEval & 0.00 & 0.00 & 0.00 \\
  & & & COGS & 0.00 & 0.00 & 0.10 \\
  & & & Entity Tracking & 1.00 & 1.00 & 1.00 \\
  & & & Explain then Translate & 0.00 & 0.00 & 1.00 \\
  & & & Instruction Tuning & 0.00 & 0.00 & 0.00 \\
  & & & Mission Impossible & 0.40 & 0.80 & 1.00 \\
  & & & Othello & 1.00 & 1.00 & 1.00 \\
  & & & Re-reading & 0.00 & 0.00 & 0.00 \\
  & & & Reasoning or Reciting & 0.00 & 0.00 & 0.40 \\
  & & & Tree Of Thoughts & 0.00 & 0.00 & 0.13 \\
  & & & VariErr-NLI & 0.00 & 0.00 & 0.40 \\
  & & & WinoDict & 0.00 & 0.40 & 0.65 \\
\cdashline{3-7}
  & & \multirow[t]{12}{*}{Detailed Hints}  & CheckEval & 0.00 & 0.00 & 0.40 \\
  & & & COGS & 0.00 & 0.60 & 0.30 \\
  & & & Entity Tracking & 0.80 & 1.00 & 1.00 \\
  & & & Explain then Translate & 0.40 & 0.80 & 0.80 \\
  & & & Instruction Tuning & 0.00 & 0.40 & 0.00 \\
  & & & Mission Impossible & 0.00 & 0.40 & 0.80 \\
  & & & Othello & 0.80 & 0.80 & 0.80 \\
  & & & Re-reading & 0.00 & 0.00 & 0.00 \\
  & & & Reasoning or Reciting & 0.00 & 0.00 & 0.44 \\
  & & & Tree Of Thoughts & 0.00 & 0.00 & 0.33 \\
  & & & VariErr-NLI & 0.00 & 0.00 & 0.80 \\
  & & & WinoDict & 0.00 & 0.00 & 0.40 \\
\bottomrule
\end{tabular}}
\caption{Detailed performance on aider + OpenAI o4-mini.}
\label{tab:detailed-aider_o4-mini}
\end{table*}

\begin{table*}
\centering
\resizebox{\linewidth}{!}{%
\renewcommand{\arraystretch}{1.3}
\footnotesize
\begin{tabular}{llllccc}
\toprule
\textbf{Agent} & \textbf{Model} & \textbf{Hint Level} & \textbf{Task} & \textbf{Final Success} & \textbf{Execution Success} & \textbf{File Recall} \\
\midrule
\multirow[t]{36}{*}{aider} & \multirow[t]{36}{*}{DeepSeek-R1} & \multirow[t]{12}{*}{No Hints}
  & CheckEval & 0.00 & 0.00 & 0.00 \\
  & & & COGS & 0.00 & 0.00 & 0.00 \\
  & & & Entity Tracking & 0.00 & 0.00 & 0.20 \\
  & & & Explain then Translate & 0.00 & 0.00 & 0.00 \\
  & & & Instruction Tuning & 0.00 & 0.00 & 0.00 \\
  & & & Mission Impossible & 0.00 & 0.00 & 0.40 \\
  & & & Othello & 0.00 & 0.00 & 0.20 \\
  & & & Re-reading & 0.00 & 0.00 & 0.13 \\
  & & & Reasoning or Reciting & 0.00 & 0.00 & 0.00 \\
  & & & Tree Of Thoughts & 0.00 & 0.00 & 0.00 \\
  & & & VariErr-NLI & 0.00 & 0.00 & 0.20 \\
  & & & WinoDict & 0.00 & 0.00 & 0.00 \\
\cdashline{3-7}
  & & \multirow[t]{12}{*}{Hints}  & CheckEval & 0.00 & 0.00 & 0.20 \\
  & & & COGS & 0.00 & 0.00 & 0.00 \\
  & & & Entity Tracking & 0.00 & 0.00 & 0.00 \\
  & & & Explain then Translate & 0.00 & 0.00 & 0.00 \\
  & & & Instruction Tuning & 0.00 & 0.00 & 0.00 \\
  & & & Mission Impossible & 0.00 & 0.00 & 0.00 \\
  & & & Othello & 0.00 & 0.00 & 0.20 \\
  & & & Re-reading & 0.00 & 0.00 & 0.00 \\
  & & & Reasoning or Reciting & 0.00 & 0.00 & 0.00 \\
  & & & Tree Of Thoughts & 0.00 & 0.00 & 0.00 \\
  & & & VariErr-NLI & 0.00 & 0.00 & 0.20 \\
  & & & WinoDict & 0.00 & 0.00 & 0.00 \\
\cdashline{3-7}
  & & \multirow[t]{12}{*}{Detailed Hints}  & CheckEval & 0.00 & 0.00 & 0.00 \\
  & & & COGS & 0.00 & 0.00 & 0.20 \\
  & & & Entity Tracking & 0.00 & 0.00 & 0.20 \\
  & & & Explain then Translate & 0.00 & 0.00 & 0.00 \\
  & & & Instruction Tuning & 0.00 & 0.00 & 0.00 \\
  & & & Mission Impossible & 0.00 & 0.00 & 0.00 \\
  & & & Othello & 0.00 & 0.00 & 0.00 \\
  & & & Re-reading & 0.00 & 0.00 & 0.00 \\
  & & & Reasoning or Reciting & 0.00 & 0.00 & 0.08 \\
  & & & Tree Of Thoughts & 0.00 & 0.00 & 0.00 \\
  & & & VariErr-NLI & 0.00 & 0.00 & 0.40 \\
  & & & WinoDict & 0.00 & 0.00 & 0.00 \\
\bottomrule
\end{tabular}}
\caption{Detailed performance on aider + DeepSeek-R1.}
\label{tab:detailed-aider_deepseek}
\end{table*}

\begin{table*}
\centering
\resizebox{\linewidth}{!}{%
\renewcommand{\arraystretch}{1.3}
\footnotesize
\begin{tabular}{llllccc}
\toprule
\textbf{Agent} & \textbf{Model} & \textbf{Hint Level} & \textbf{Task} & \textbf{Final Success} & \textbf{Execution Success} & \textbf{File Recall} \\
\midrule
\multirow[t]{36}{*}{OpenHands} & \multirow[t]{36}{*}{Claude 4 Sonnet} & \multirow[t]{12}{*}{No Hints}
  & CheckEval & 0.00 & 0.80 & 0.40 \\
  & & & COGS & 1.00 & 1.00 & 0.60 \\
  & & & Entity Tracking & 0.00 & 0.00 & 1.00 \\
  & & & Explain then Translate & 0.60 & 1.00 & 1.00 \\
  & & & Instruction Tuning & 0.40 & 0.40 & 0.00 \\
  & & & Mission Impossible & 0.00 & 0.60 & 0.80 \\
  & & & Othello & 1.00 & 1.00 & 1.00 \\
  & & & Re-reading & 0.00 & 0.80 & 1.00 \\
  & & & Reasoning or Reciting & 0.00 & 0.60 & 0.40 \\
  & & & Tree Of Thoughts & 0.40 & 1.00 & 0.67 \\
  & & & VariErr-NLI & 0.00 & 0.00 & 1.00 \\
  & & & WinoDict & 0.60 & 1.00 & 0.25 \\
\cdashline{3-7}
  & & \multirow[t]{12}{*}{Hints}  & CheckEval & 0.00 & 0.20 & 0.80 \\
  & & & COGS & 0.60 & 0.60 & 0.50 \\
  & & & Entity Tracking & 0.80 & 0.80 & 1.00 \\
  & & & Explain then Translate & 1.00 & 1.00 & 1.00 \\
  & & & Instruction Tuning & 0.60 & 0.80 & 0.00 \\
  & & & Mission Impossible & 0.80 & 1.00 & 1.00 \\
  & & & Othello & 0.80 & 0.80 & 1.00 \\
  & & & Re-reading & 0.00 & 0.80 & 0.93 \\
  & & & Reasoning or Reciting & 0.00 & 0.60 & 0.40 \\
  & & & Tree Of Thoughts & 0.40 & 0.80 & 0.67 \\
  & & & VariErr-NLI & 0.00 & 0.20 & 1.00 \\
  & & & WinoDict & 0.20 & 0.60 & 0.65 \\
\cdashline{3-7}
  & & \multirow[t]{12}{*}{Detailed Hints}  & CheckEval & 0.40 & 0.80 & 1.00 \\
  & & & COGS & 0.80 & 0.80 & 1.00 \\
  & & & Entity Tracking & 0.20 & 1.00 & 1.00 \\
  & & & Explain then Translate & 1.00 & 1.00 & 1.00 \\
  & & & Instruction Tuning & 0.20 & 0.20 & 0.00 \\
  & & & Mission Impossible & 0.60 & 1.00 & 1.00 \\
  & & & Othello & 1.00 & 1.00 & 1.00 \\
  & & & Re-reading & 0.00 & 1.00 & 0.87 \\
  & & & Reasoning or Reciting & 0.00 & 1.00 & 0.40 \\
  & & & Tree Of Thoughts & 0.80 & 1.00 & 0.67 \\
  & & & VariErr-NLI & 0.00 & 0.00 & 1.00 \\
  & & & WinoDict & 0.20 & 0.60 & 0.75 \\
\bottomrule
\end{tabular}}
\caption{Detailed performance on OpenHands + Claude 4 Sonnet.}
\label{tab:detailed-openhands_claude_4}
\end{table*}

\begin{table*}
\centering
\resizebox{\linewidth}{!}{%
\renewcommand{\arraystretch}{1.3}
\footnotesize
\begin{tabular}{llllccc}
\toprule
\textbf{Agent} & \textbf{Model} & \textbf{Hint Level} & \textbf{Task} & \textbf{Final Success} & \textbf{Execution Success} & \textbf{File Recall} \\
\midrule
\multirow[t]{36}{*}{OpenHands} & \multirow[t]{36}{*}{Claude 3.7 Sonnet} & \multirow[t]{12}{*}{No Hints}
  & CheckEval & 0.00 & 0.00 & 0.50 \\
  & & & COGS & 0.80 & 1.00 & 0.50 \\
  & & & Entity Tracking & 0.00 & 0.00 & 1.00 \\
  & & & Explain then Translate & 1.00 & 1.00 & 1.00 \\
  & & & Instruction Tuning & 0.40 & 0.80 & 0.00 \\
  & & & Mission Impossible & 0.00 & 0.00 & 0.50 \\
  & & & Othello & 1.00 & 1.00 & 1.00 \\
  & & & Re-reading & 0.00 & 0.40 & 0.73 \\
  & & & Reasoning or Reciting & 0.00 & 0.00 & 0.36 \\
  & & & Tree Of Thoughts & 0.00 & 0.40 & 0.67 \\
  & & & VariErr-NLI & 0.00 & 0.40 & 1.00 \\
  & & & WinoDict & 0.00 & 0.40 & 0.25 \\
\cdashline{3-7}
  & & \multirow[t]{12}{*}{Hints}  & CheckEval & 0.40 & 0.80 & 0.80 \\
  & & & COGS & 1.00 & 1.00 & 1.00 \\
  & & & Entity Tracking & 1.00 & 1.00 & 1.00 \\
  & & & Explain then Translate & 0.80 & 1.00 & 1.00 \\
  & & & Instruction Tuning & 0.40 & 0.40 & 0.00 \\
  & & & Mission Impossible & 0.00 & 0.60 & 1.00 \\
  & & & Othello & 1.00 & 1.00 & 1.00 \\
  & & & Re-reading & 0.00 & 0.00 & 0.67 \\
  & & & Reasoning or Reciting & 0.00 & 0.00 & 0.40 \\
  & & & Tree Of Thoughts & 0.20 & 0.60 & 0.67 \\
  & & & VariErr-NLI & 0.00 & 0.00 & 1.00 \\
  & & & WinoDict & 0.00 & 0.00 & 0.70 \\
\cdashline{3-7}
  & & \multirow[t]{12}{*}{Detailed Hints}  & CheckEval & 0.00 & 0.40 & 1.00 \\
  & & & COGS & 1.00 & 1.00 & 1.00 \\
  & & & Entity Tracking & 1.00 & 1.00 & 1.00 \\
  & & & Explain then Translate & 1.00 & 1.00 & 1.00 \\
  & & & Instruction Tuning & 0.00 & 0.60 & 0.00 \\
  & & & Mission Impossible & 1.00 & 1.00 & 1.00 \\
  & & & Othello & 0.00 & 1.00 & 1.00 \\
  & & & Re-reading & 0.00 & 0.00 & 0.67 \\
  & & & Reasoning or Reciting & 0.00 & 0.00 & 0.40 \\
  & & & Tree Of Thoughts & 0.20 & 0.20 & 0.60 \\
  & & & VariErr-NLI & 0.00 & 0.00 & 1.00 \\
  & & & WinoDict & 0.00 & 0.00 & 0.75 \\
\bottomrule
\end{tabular}}
\caption{Detailed performance on OpenHands + Claude 3.7 Sonnet.}
\label{tab:detailed-openhands_claude}
\end{table*}

\begin{table*}
\centering
\resizebox{\linewidth}{!}{%
\renewcommand{\arraystretch}{1.3}
\footnotesize
\begin{tabular}{llllccc}
\toprule
\textbf{Agent} & \textbf{Model} & \textbf{Hint Level} & \textbf{Task} & \textbf{Final Success} & \textbf{Execution Success} & \textbf{File Recall} \\
\midrule
\multirow[t]{36}{*}{OpenHands} & \multirow[t]{36}{*}{OpenAI GPT-5} & \multirow[t]{12}{*}{No Hints}
  & CheckEval & 0.00 & 1.00 & 0.50 \\
  & & & COGS & 0.80 & 0.80 & 0.50 \\
  & & & Entity Tracking & 0.00 & 0.00 & 1.00 \\
  & & & Explain then Translate & 0.60 & 0.80 & 1.00 \\
  & & & Instruction Tuning & 0.00 & 0.00 & 0.00 \\
  & & & Mission Impossible & 0.20 & 0.60 & 0.80 \\
  & & & Othello & 1.00 & 1.00 & 1.00 \\
  & & & Re-reading & 0.00 & 0.60 & 0.93 \\
  & & & Reasoning or Reciting & 0.00 & 0.80 & 0.40 \\
  & & & Tree Of Thoughts & 0.20 & 1.00 & 0.67 \\
  & & & VariErr-NLI & 0.00 & 0.40 & 1.00 \\
  & & & WinoDict & 0.40 & 0.40 & 0.30 \\
\cdashline{3-7}
  & & \multirow[t]{12}{*}{Hints}  & CheckEval & 0.00 & 0.20 & 1.00 \\
  & & & COGS & 1.00 & 1.00 & 0.50 \\
  & & & Entity Tracking & 0.40 & 0.60 & 1.00 \\
  & & & Explain then Translate & 1.00 & 1.00 & 1.00 \\
  & & & Instruction Tuning & 0.00 & 0.00 & 0.00 \\
  & & & Mission Impossible & 0.00 & 0.20 & 1.00 \\
  & & & Othello & 0.60 & 0.60 & 1.00 \\
  & & & Re-reading & 0.00 & 0.80 & 0.93 \\
  & & & Reasoning or Reciting & 0.00 & 1.00 & 0.40 \\
  & & & Tree Of Thoughts & 0.80 & 1.00 & 0.67 \\
  & & & VariErr-NLI & 0.00 & 0.40 & 1.00 \\
  & & & WinoDict & 0.60 & 0.80 & 0.75 \\
\cdashline{3-7}
  & & \multirow[t]{12}{*}{Detailed Hints}  & CheckEval & 0.40 & 0.60 & 0.90 \\
  & & & COGS & 1.00 & 1.00 & 0.50 \\
  & & & Entity Tracking & 0.40 & 0.40 & 1.00 \\
  & & & Explain then Translate & 1.00 & 1.00 & 1.00 \\
  & & & Instruction Tuning & 0.00 & 0.00 & 0.00 \\
  & & & Mission Impossible & 0.20 & 0.40 & 1.00 \\
  & & & Othello & 1.00 & 1.00 & 1.00 \\
  & & & Re-reading & 0.00 & 1.00 & 1.00 \\
  & & & Reasoning or Reciting & 0.00 & 1.00 & 0.40 \\
  & & & Tree Of Thoughts & 0.80 & 0.80 & 0.67 \\
  & & & VariErr-NLI & 0.00 & 0.20 & 1.00 \\
  & & & WinoDict & 0.40 & 1.00 & 0.75 \\
\bottomrule
\end{tabular}}
\caption{Detailed performance on OpenHands + OpenAI GPT-5.}
\label{tab:detailed-openhands_gpt5}
\end{table*}

\begin{table*}
\centering
\resizebox{\linewidth}{!}{%
\renewcommand{\arraystretch}{1.3}
\footnotesize
\begin{tabular}{llllccc}
\toprule
\textbf{Agent} & \textbf{Model} & \textbf{Hint Level} & \textbf{Task} & \textbf{Final Success} & \textbf{Execution Success} & \textbf{File Recall} \\
\midrule
\multirow[t]{36}{*}{OpenHands} & \multirow[t]{36}{*}{OpenAI o1} & \multirow[t]{12}{*}{No Hints}
  & CheckEval & 0.00 & 1.00 & 0.50 \\
  & & & COGS & 0.00 & 0.00 & 0.40 \\
  & & & Entity Tracking & 0.00 & 0.40 & 1.00 \\
  & & & Explain then Translate & 0.00 & 1.00 & 0.80 \\
  & & & Instruction Tuning & 0.00 & 0.00 & 0.00 \\
  & & & Mission Impossible & 0.00 & 0.00 & 0.50 \\
  & & & Othello & 0.00 & 0.00 & 0.80 \\
  & & & Re-reading & 0.00 & 0.40 & 0.67 \\
  & & & Reasoning or Reciting & 0.00 & 0.00 & 0.24 \\
  & & & Tree Of Thoughts & 0.00 & 0.40 & 0.53 \\
  & & & VariErr-NLI & 0.00 & 0.00 & 0.80 \\
  & & & WinoDict & 0.00 & 0.80 & 0.25 \\
\cdashline{3-7}
  & & \multirow[t]{12}{*}{Hints}  & CheckEval & 0.00 & 1.00 & 0.70 \\
  & & & COGS & 0.00 & 0.00 & 0.50 \\
  & & & Entity Tracking & 1.00 & 1.00 & 1.00 \\
  & & & Explain then Translate & 0.00 & 0.40 & 1.00 \\
  & & & Instruction Tuning & 0.00 & 0.00 & 0.00 \\
  & & & Mission Impossible & 0.00 & 0.60 & 0.50 \\
  & & & Othello & 0.00 & 0.00 & 1.00 \\
  & & & Re-reading & 0.00 & 0.40 & 0.73 \\
  & & & Reasoning or Reciting & 0.00 & 0.00 & 0.24 \\
  & & & Tree Of Thoughts & 0.00 & 0.60 & 0.60 \\
  & & & VariErr-NLI & 0.00 & 0.00 & 0.70 \\
  & & & WinoDict & 0.00 & 0.00 & 0.65 \\
\cdashline{3-7}
  & & \multirow[t]{12}{*}{Detailed Hints}  & CheckEval & 0.00 & 0.00 & 0.80 \\
  & & & COGS & 0.00 & 1.00 & 0.90 \\
  & & & Entity Tracking & 0.00 & 0.60 & 0.80 \\
  & & & Explain then Translate & 0.40 & 0.60 & 0.60 \\
  & & & Instruction Tuning & 0.00 & 0.40 & 0.00 \\
  & & & Mission Impossible & 0.00 & 0.00 & 0.80 \\
  & & & Othello & 0.00 & 0.40 & 0.90 \\
  & & & Re-reading & 0.00 & 0.40 & 0.93 \\
  & & & Reasoning or Reciting & 0.00 & 0.00 & 0.40 \\
  & & & Tree Of Thoughts & 0.00 & 1.00 & 0.60 \\
  & & & VariErr-NLI & 0.00 & 0.00 & 0.70 \\
  & & & WinoDict & 0.00 & 0.20 & 0.45 \\
\bottomrule
\end{tabular}}
\caption{Detailed performance on OpenHands + OpenAI o1.}
\label{tab:detailed-openhands_o1}
\end{table*}

\begin{table*}
\centering
\resizebox{\linewidth}{!}{%
\renewcommand{\arraystretch}{1.3}
\footnotesize
\begin{tabular}{llllccc}
\toprule
\textbf{Agent} & \textbf{Model} & \textbf{Hint Level} & \textbf{Task} & \textbf{Final Success} & \textbf{Execution Success} & \textbf{File Recall} \\
\midrule
\multirow[t]{36}{*}{OpenHands} & \multirow[t]{36}{*}{OpenAI o4-mini} & \multirow[t]{12}{*}{No Hints}
  & CheckEval & 0.00 & 0.40 & 0.50 \\
  & & & COGS & 0.20 & 0.20 & 0.50 \\
  & & & Entity Tracking & 0.40 & 1.00 & 1.00 \\
  & & & Explain then Translate & 0.00 & 0.60 & 1.00 \\
  & & & Instruction Tuning & 0.00 & 0.00 & 0.00 \\
  & & & Mission Impossible & 0.00 & 0.00 & 0.10 \\
  & & & Othello & 0.40 & 0.40 & 0.90 \\
  & & & Re-reading & 0.00 & 0.60 & 1.00 \\
  & & & Reasoning or Reciting & 0.00 & 0.40 & 0.40 \\
  & & & Tree of Thoughts & 0.00 & 0.60 & 0.00 \\
  & & & VariErr-NLI & 0.00 & 0.00 & 0.80 \\
  & & & WinoDict & 0.00 & 0.20 & 0.25 \\
\cdashline{3-7}
  & & \multirow[t]{12}{*}{Hints}  & CheckEval & 0.00 & 0.00 & 0.50 \\
  & & & COGS & 0.80 & 0.80 & 0.50 \\
  & & & Entity Tracking & 0.80 & 1.00 & 1.00 \\
  & & & Explain then Translate & 0.00 & 0.20 & 1.00 \\
  & & & Instruction Tuning & 0.00 & 0.00 & 0.00 \\
  & & & Mission Impossible & 0.00 & 0.80 & 0.70 \\
  & & & Othello & 0.60 & 0.60 & 0.90 \\
  & & & Re-reading & 0.40 & 1.00 & 0.93 \\
  & & & Reasoning or Reciting & 0.00 & 0.00 & 0.32 \\
  & & & Tree of Thoughts & 0.00 & 0.00 & 0.00 \\
  & & & VariErr-NLI & 0.00 & 0.00 & 0.90 \\
  & & & WinoDict & 0.00 & 0.00 & 0.70 \\
\cdashline{3-7}
  & & \multirow[t]{12}{*}{Detailed Hints}  & CheckEval & 0.00 & 0.40 & 0.60 \\
  & & & COGS & 0.60 & 0.60 & 0.50 \\
  & & & Entity Tracking & 0.00 & 0.80 & 0.80 \\
  & & & Explain then Translate & 0.00 & 0.80 & 1.00 \\
  & & & Instruction Tuning & 0.00 & 0.00 & 0.00 \\
  & & & Mission Impossible & 0.00 & 0.60 & 1.00 \\
  & & & Othello & 1.00 & 1.00 & 1.00 \\
  & & & Re-reading & 0.00 & 1.00 & 1.00 \\
  & & & Reasoning or Reciting & 0.00 & 0.60 & 0.40 \\
  & & & Tree of Thoughts & 0.00 & 0.00 & 0.00 \\
  & & & VariErr-NLI & 0.00 & 0.00 & 0.60 \\
  & & & WinoDict & 0.00 & 0.00 & 0.60 \\
\bottomrule
\end{tabular}}
\caption{Detailed performance on OpenHands + OpenAI o4-mini.}
\label{tab:detailed-openhands_o4-mini}
\end{table*}

\begin{table*}
\centering
\resizebox{\linewidth}{!}{%
\renewcommand{\arraystretch}{1.3}
\footnotesize
\begin{tabular}{llllccc}
\toprule
\textbf{Agent} & \textbf{Model} & \textbf{Hint Level} & \textbf{Task} & \textbf{Final Success} & \textbf{Execution Success} & \textbf{File Recall} \\
\midrule
\multirow[t]{36}{*}{OpenHands} & \multirow[t]{36}{*}{DeepSeek-R1} & \multirow[t]{12}{*}{No Hints}
  & CheckEval & 0.00 & 0.00 & 0.40 \\
  & & & COGS & 0.00 & 0.40 & 0.40 \\
  & & & Entity Tracking & 0.00 & 0.00 & 0.60 \\
  & & & Explain then Translate & 0.00 & 0.40 & 1.00 \\
  & & & Instruction Tuning & 0.00 & 0.40 & 0.00 \\
  & & & Mission Impossible & 0.00 & 0.00 & 0.30 \\
  & & & Othello & 0.00 & 0.40 & 1.00 \\
  & & & Re-reading & 0.00 & 0.00 & 0.40 \\
  & & & Reasoning or Reciting & 0.00 & 0.00 & 0.28 \\
  & & & Tree Of Thoughts & 0.00 & 0.40 & 0.67 \\
  & & & VariErr-NLI & 0.00 & 0.00 & 0.80 \\
  & & & WinoDict & 0.00 & 0.00 & 0.15 \\
\cdashline{3-7}
  & & \multirow[t]{12}{*}{Hints}  & CheckEval & 0.00 & 0.20 & 0.50 \\
  & & & COGS & 0.00 & 0.00 & 0.50 \\
  & & & Entity Tracking & 0.00 & 0.00 & 0.80 \\
  & & & Explain then Translate & 0.00 & 0.60 & 0.80 \\
  & & & Instruction Tuning & 0.00 & 0.00 & 0.00 \\
  & & & Mission Impossible & 0.00 & 0.00 & 0.60 \\
  & & & Othello & 0.00 & 0.40 & 0.80 \\
  & & & Re-reading & 0.00 & 0.00 & 0.60 \\
  & & & Reasoning or Reciting & 0.00 & 0.00 & 0.44 \\
  & & & Tree Of Thoughts & 0.00 & 0.00 & 0.47 \\
  & & & VariErr-NLI & 0.00 & 0.40 & 1.00 \\
  & & & WinoDict & 0.00 & 0.00 & 0.30 \\
\cdashline{3-7}
  & & \multirow[t]{12}{*}{Detailed Hints}  & CheckEval & 0.00 & 0.00 & 0.60 \\
  & & & COGS & 0.00 & 0.20 & 0.50 \\
  & & & Entity Tracking & 0.00 & 0.60 & 0.80 \\
  & & & Explain then Translate & 0.80 & 1.00 & 1.00 \\
  & & & Instruction Tuning & 0.00 & 0.00 & 0.00 \\
  & & & Mission Impossible & 0.00 & 0.00 & 1.00 \\
  & & & Othello & 0.40 & 0.40 & 0.80 \\
  & & & Re-reading & 0.00 & 0.00 & 0.93 \\
  & & & Reasoning or Reciting & 0.00 & 0.00 & 0.32 \\
  & & & Tree Of Thoughts & 0.00 & 0.00 & 0.53 \\
  & & & VariErr-NLI & 0.00 & 0.00 & 1.00 \\
  & & & WinoDict & 0.00 & 0.40 & 0.15 \\
\bottomrule
\end{tabular}}
\caption{Detailed performance on OpenHands + DeepSeek-R1.}
\label{tab:detailed-openhands_deepseek-r1}
\end{table*}

\begin{table*}
\centering
\resizebox{\linewidth}{!}{%
\renewcommand{\arraystretch}{1.3}
\footnotesize
\begin{tabular}{llllccc}
\toprule
\textbf{Agent} & \textbf{Model} & \textbf{Hint Level} & \textbf{Task} & \textbf{Final Success} & \textbf{Execution Success} & \textbf{File Recall} \\
\midrule
\multirow[t]{36}{*}{aider} & \multirow[t]{36}{*}{Claude 4.5 Opus} & \multirow[t]{12}{*}{No Hints}
  & CheckEval & 0.00 & 0.80 & 0.80 \\
  & & & COGS & 0.00 & 0.20 & 0.50 \\
  & & & Entity Tracking & 1.00 & 1.00 & 1.00 \\
  & & & Explain then Translate & 0.00 & 1.00 & 1.00 \\
  & & & Implicit Instructions & 0.00 & 0.00 & 0.50 \\
  & & & Mission Impossible & 0.40 & 0.80 & 1.00 \\
  & & & Othello & 1.00 & 1.00 & 1.00 \\
  & & & Re-reading & 0.00 & 0.20 & 0.67 \\
  & & & Reasoning or Reciting & 0.00 & 0.00 & 0.60 \\
  & & & Tree Of Thoughts & 0.40 & 0.40 & 0.67 \\
  & & & VariErr-NLI & 0.00 & 0.00 & 1.00 \\
  & & & WinoDict & 0.00 & 0.20 & 0.75 \\
\cdashline{3-7}
  & & \multirow[t]{12}{*}{Hints}  & CheckEval & 0.00 & 0.80 & 0.80 \\
  & & & COGS & 0.00 & 0.00 & 0.30 \\
  & & & Entity Tracking & 0.60 & 0.60 & 0.60 \\
  & & & Explain then Translate & 0.00 & 0.60 & 0.60 \\
  & & & Implicit Instructions & 0.00 & 0.00 & 0.30 \\
  & & & Mission Impossible & 0.20 & 0.80 & 0.80 \\
  & & & Othello & 0.80 & 0.80 & 0.80 \\
  & & & Re-reading & 0.00 & 0.00 & 0.53 \\
  & & & Reasoning or Reciting & 0.00 & 0.00 & 0.48 \\
  & & & Tree Of Thoughts & 0.00 & 0.00 & 0.40 \\
  & & & VariErr-NLI & 0.00 & 0.00 & 0.60 \\
  & & & WinoDict & 0.00 & 0.00 & 0.45 \\
\cdashline{3-7}
  & & \multirow[t]{12}{*}{Detailed Hints}  & CheckEval & 0.00 & 0.80 & 1.00 \\
  & & & COGS & 0.00 & 0.00 & 0.90 \\
  & & & Entity Tracking & 0.40 & 0.40 & 1.00 \\
  & & & Explain then Translate & 1.00 & 1.00 & 1.00 \\
  & & & Implicit Instructions & 0.00 & 0.00 & 0.50 \\
  & & & Mission Impossible & 0.20 & 1.00 & 1.00 \\
  & & & Othello & 1.00 & 1.00 & 1.00 \\
  & & & Re-reading & 0.00 & 0.20 & 0.67 \\
  & & & Reasoning or Reciting & 0.00 & 0.00 & 0.60 \\
  & & & Tree Of Thoughts & 0.00 & 0.00 & 0.87 \\
  & & & VariErr-NLI & 0.00 & 0.00 & 1.00 \\
  & & & WinoDict & 0.00 & 0.00 & 0.75 \\
\bottomrule
\end{tabular}}
\caption{Detailed performance on aider + Claude 4.5 Opus.}
\label{tab:detailed-aider_claude_4.5_opus}
\end{table*}

\begin{table*}
\centering
\resizebox{\linewidth}{!}{%
\renewcommand{\arraystretch}{1.3}
\footnotesize
\begin{tabular}{llllccc}
\toprule
\textbf{Agent} & \textbf{Model} & \textbf{Hint Level} & \textbf{Task} & \textbf{Final Success} & \textbf{Execution Success} & \textbf{File Recall} \\
\midrule
\multirow[t]{36}{*}{OpenHands} & \multirow[t]{36}{*}{Claude 4.5 Opus} & \multirow[t]{12}{*}{No Hints}
  & CheckEval & 0.00 & 1.00 & 0.50 \\
  & & & COGS & 1.00 & 1.00 & 0.50 \\
  & & & Entity Tracking & 0.00 & 0.00 & 1.00 \\
  & & & Explain then Translate & 0.60 & 1.00 & 1.00 \\
  & & & Instruction Tuning & 0.60 & 0.00 & 0.00 \\
  & & & Mission Impossible & 0.20 & 0.80 & 1.00 \\
  & & & Othello & 1.00 & 1.00 & 1.00 \\
  & & & Re-reading & 0.20 & 1.00 & 0.67 \\
  & & & Reasoning or Reciting & 0.00 & 0.40 & 0.40 \\
  & & & Tree Of Thoughts & 1.00 & 1.00 & 0.67 \\
  & & & VariErr-NLI & 0.00 & 0.20 & 1.00 \\
  & & & WinoDict & 0.40 & 1.00 & 0.25 \\
\cdashline{3-7}
  & & \multirow[t]{12}{*}{Hints}  & CheckEval & 0.40 & 1.00 & 1.00 \\
  & & & COGS & 0.80 & 0.80 & 0.50 \\
  & & & Entity Tracking & 1.00 & 1.00 & 1.00 \\
  & & & Explain then Translate & 1.00 & 1.00 & 1.00 \\
  & & & Instruction Tuning & 0.40 & 0.00 & 0.00 \\
  & & & Mission Impossible & 0.80 & 1.00 & 1.00 \\
  & & & Othello & 0.80 & 1.00 & 1.00 \\
  & & & Re-reading & 0.40 & 1.00 & 0.67 \\
  & & & Reasoning or Reciting & 0.00 & 1.00 & 0.40 \\
  & & & Tree Of Thoughts & 1.00 & 1.00 & 0.67 \\
  & & & VariErr-NLI & 0.00 & 0.00 & 1.00 \\
  & & & WinoDict & 0.80 & 0.80 & 0.75 \\
\cdashline{3-7}
  & & \multirow[t]{12}{*}{Detailed Hints}  & CheckEval & 0.00 & 1.00 & 0.50 \\
  & & & COGS & 1.00 & 1.00 & 1.00 \\
  & & & Entity Tracking & 0.00 & 1.00 & 1.00 \\
  & & & Explain then Translate & 1.00 & 1.00 & 1.00 \\
  & & & Instruction Tuning & 0.00 & 0.00 & 0.00 \\
  & & & Mission Impossible & 0.60 & 1.00 & 1.00 \\
  & & & Othello & 0.80 & 1.00 & 1.00 \\
  & & & Re-reading & 1.00 & 1.00 & 0.73 \\
  & & & Reasoning or Reciting & 0.00 & 1.00 & 0.40 \\
  & & & Tree Of Thoughts & 1.00 & 1.00 & 0.67 \\
  & & & VariErr-NLI & 0.00 & 0.00 & 1.00 \\
  & & & WinoDict & 0.60 & 0.80 & 0.75 \\
\bottomrule
\end{tabular}}
\caption{Detailed performance on OpenHands + Claude 4.5 Opus.}
\label{tab:detailed-openhands_claude_4.5_opus}
\end{table*}

\section{Additional qualitative observations}
\label{app:qualitative-observations}
\paragraph{Some agent edits have no practical effect} Although stronger agents more often write executable code, sometimes the actual implementation has no effect on the output. For example, in the Mission Impossible task, both OpenHands + \{Claude 4 Sonnet,  GPT-5\} incorrectly used the \texttt{ParentedTree} class in the \texttt{nltk} library. While this code raised a \texttt{ValueError}, the agent's implementation used a \texttt{try-except} block, returning the parse tree from the original paper as a fallback value, meaning the script still executed. In another instance, OpenHands + GPT-5 incorrectly tried to access the content returned by a method in the \texttt{radon} library. The code logic meant that if no function was found using this method, a default value of 0.0 was returned. This in effect meant that the final numerical results were identical to the original paper's experiments. These observations reaffirm the importance of rigorous verification before deploying these systems in the real world.

\paragraph{Detailed observations on over-editing} As mentioned in \Cref{sec:error-analysis}, despite the constraints specified in our task instructions for the agents, we occassionally observed agents making unnecessary additional code edits. For example, for the Re-reading task, OpenHands + Claude Sonnet 4 unnecessarily modified an additional metadata field in one of the \texttt{.yml} files, which is used as part of the input in one of the experimental settings. Similarly, in the VariErr-NLI task OpenHands + GPT-5 unnecessarily modified an output file path required for obtaining the final scores, resulting in the evaluation scripts being unable to access the results of the agent's implementation. Given that scientific work relies on rigor and reproducibility, deviations like these from the specified instructions are problematic. This highlights the need to design agents which conform exactly to the requirements given, without introducing additional unrequested changes.

\section{Additional Results with Claude 4.5 Opus}
\label{appendix:additional_results_claude_4.5}
After submission, we additionally evaluated Claude 4.5 Opus, a recently released frontier model, with both aider and OpenHands. Even with this stronger model, final success remains below 45\%: OpenHands + Claude 4.5 Opus reaches 42\% and aider + Claude 4.5 Opus reaches 23\% (\Cref{fig:main_results_appendix}), indicating that \benchmark{} is not saturated by current frontier models. Still, the signal from our benchmark is clear: newer models consistently improve over previously evaluated models (33\% and 19\%, respectively, with Claude 4 Sonnet), and execution success with OpenHands rises to 70\%. Hints further boost performance (\Cref{fig:hint_results_appendix}), with OpenHands + Claude 4.5 Opus reaching 62\% final success, consistent with the pattern observed in the main experiments.

The distribution of execution errors (\Cref{fig:explicit_error_types_appendix}) follows a similar pattern to other models, with FileNotFoundError and ValueError accounting for the majority of failures, showing that the primary failure modes remain consistent even with stronger models.

\section{License Information}

The codebase portion of \benchmark{} is constructed from public repositories---details of the licenses for each task are provided in \Cref{tab:task-license-info}. When the codebase did not contain any license information, we reached out to the authors for more information and used their suggestion (when we did not receive a response, we made an educated guess that the repository will be associated with a permissive license given that the paper was written by authors with primarily academic affiliations, and from the public availability of their codebase). We release our data and code under a dual license (MIT and Apache 2.0), given the mixed license of the repositories included in the full benchmark suite.

\end{document}